%% file: main.tex
\documentclass{article}

\usepackage[final,nonatbib]{neurips_2022}

\usepackage[utf8]{inputenc} % allow utf-8 input
\usepackage[T1]{fontenc}    % use 8-bit T1 fonts
\usepackage[colorlinks,pagebackref]{hyperref}       % hyperlinks
\usepackage{url}            % simple URL typesetting
\usepackage{booktabs}       % professional-quality tables
\usepackage{amsfonts}       % blackboard math symbols
\usepackage{nicefrac}       % compact symbols for 1/2, etc.
\usepackage{microtype}      % microtypography
\usepackage{float}
\usepackage[dvipsnames]{xcolor}
\usepackage{graphicx}
\usepackage{dsfont}
\usepackage{amsmath,amssymb}
\usepackage{xspace}
\usepackage{wrapfig}
\usepackage{multirow}
\usepackage{pifont}
\usepackage{lipsum}
\usepackage{rotating}
\usepackage{bm}
\usepackage{adjustbox}

\definecolor{mygreen}{rgb}{0.9,0.65,0.8}
\definecolor{mylightorange}{rgb}{0.9,0.72,0.55}
\definecolor{mycyan}{rgb}{0.7,0.9,1}
\definecolor{mygray}{rgb}{0.92,0.92,0.92}

\makeatletter
\DeclareRobustCommand\onedot{\futurelet\@let@token\@onedot}
\def\@onedot{\ifx\@let@token.\else.\null\fi\xspace}

\def\eg{\emph{e.g}\onedot} 
\def\ie{\emph{i.e}\onedot} 
 
\def\etc{\emph{etc}\onedot}

\def\etal{\emph{et al}\onedot}
\makeatother

\newcommand{\mypara}[1]{\vspace{0mm}\noindent\textbf{#1}}

\newcommand{\be}{\mathbf{e}}
\newcommand{\bo}{\mathbf{o}}
\newcommand{\bx}{\mathbf{x}}
\newcommand{\br}{\mathbf{r}}
\newcommand{\bq}{\mathbf{q}}
\newcommand{\bk}{\mathbf{k}}
\newcommand{\bv}{\mathbf{v}}
\newcommand{\bz}{\mathbf{z}}
\newcommand{\PE}{\text{PE}}
\newcommand{\mcE}{\mathcal{E}}
\newcommand{\mcF}{\mathcal{F}}
\newcommand{\mcC}{\mathcal{C}}
\newcommand{\mcR}{\mathcal{R}}
\newcommand{\mcP}{\mathcal{P}}
\newcommand{\mcB}{\mathcal{B}}
\newcommand{\mcL}{\mathcal{L}}
\newcommand{\mcG}{\mathcal{G}}

\newcommand{\modelname}{VideoWhisperer}
\title{Grounded Video Situation Recognition}

\author{%
Zeeshan Khan \hspace{0.5cm} C. V. Jawahar \hspace{0.5cm} Makarand Tapaswi\\
CVIT, IIIT Hyderabad \\
{\small \url{https://zeeshank95.github.io/grvidsitu}}
}

\begin{document}

\maketitle

\input{sections/0_abstract}
\input{sections/1_introduction}

\input{sections/2_related}

\input{sections/3_method}

\input{sections/4_experiments}

\input{sections/5_conclusion}

{\small
\bibliographystyle{ieee_fullname}
\bibliography{longstrings,references}
}

\appendix

\newpage
\input{sections/supplementary}

\end{document}

%% file: sections/0_abstract.tex
\begin{abstract}

Dense video understanding requires answering several questions such as \emph{who is doing what to whom, with what, how, why, and where}.
Recently, Video Situation Recognition (VidSitu) is framed as a task for structured prediction of multiple events, their relationships, and actions and various verb-role pairs attached to descriptive entities.
% Video semantic role labelling (VidSRL)~\cite{sadhu2021vidsitu} is a dense video understanding task that requires recognizing multiple events and the roles and entities involved in them captured as free-form captions.
This task poses several challenges in identifying, disambiguating, and co-referencing entities across multiple verb-role pairs, but also faces some challenges of evaluation.
% VidSitu poses several challenges including disambiguating the roles for all possible entities, generating good descriptive and distinctive captions to uniquely represent each role, and co-referencing all the entities throughout the video.
In this work, we propose the addition of spatio-temporal grounding as an essential component of the structured prediction task in a weakly supervised setting,
and present a novel three stage Transformer model, \emph{\modelname{}}, that is empowered to make joint predictions.
In stage one, we learn contextualised embeddings for video features in parallel with key objects that appear in the video clips to enable fine-grained spatio-temporal reasoning.
% , and learn their contextualised embeddings using self attention enabling both fine-grained spatial and temporal reasoning.
The second stage sees verb-role queries attend and pool information from object embeddings, localising \emph{answers} to questions posed about the action.
% event aware cross attention, where role queries attend to contextualised event-aware object embeddings and search for the best objects that represents each role.
The final stage generates these answers as captions to describe each verb-role pair present in the video.
% for each role that considers the output of the role feaures.
Our model operates on a group of events (clips) simultaneously and predicts verbs, verb-role pairs, their nouns, and their grounding on-the-fly.
When evaluated on a grounding-augmented version of the VidSitu dataset, we observe a large improvement in entity captioning accuracy, as well as the ability to localize verb-roles without grounding annotations at training time.
% In addition to improving captioning accuracy our model allows to ground each role in the spatio-temporal domain by exploiting the role-object cross-attention in a weakly-supervised manner.
% Finally, different from previous works that require the verb to model SRL, we propose to predict verb-role pairs on the fly and design role-queries that are contextualised by event embeddings, circumventing the need for event verbs to model SRL, and enabling end-to-end grounded situation recognition in videos (GSRLV).
% We evaluate GSRLV on the VidSitu dataset~\cite{sadhu2021vidsitu}, showing large performance improvement.
% , and provide baselines for future work.
\end{abstract}

%% file: sections/1_introduction.tex
\section{Introduction}

At the end of \emph{The Dark Knight}, we see a short intense sequence%
% \footnote{Spoiler warning! The first 10 seconds of \url{https://www.youtube.com/watch?v=I6FfPTg1iic}.}
that involves Harvey Dent toss a coin while holding a gun followed by sudden action.
Holistic understanding of such a video sequence, especially one that involves multiple people, requires predicting more than the action label (\emph{what} verb).
For example, we may wish to answer questions such as \emph{who} performed the action (agent), \emph{why} they are doing it (purpose / goal), \emph{how} are they doing it (manner), \emph{where} are they doing it (location), and even \emph{what happens after} (multi-event understanding).
% Humans can continuously process visual signals with high cognition ability at the finest details. We are able to coherently perceive complicated situations by quickly recognizing the environment, agents/objects, their interactions, and the role they play.
% Machines on the other hand works best with independent uni-dimensional tasks with limited coherence, such as understanding actions, detecting objects etc. In order to embed high cognition abilities in machines, it would require understanding situations with both local and global level dependencies. Videos are integral part of our lives, they contain rich information of the physical world at multiple levels of abstraction.
While humans are able to perceive the situation and are good at answering such questions, many works often focus on building tools for doing single tasks, \eg~predicting actions~\cite{feichtenhofer2019slowfast} or detecting objects~\cite{anderson2018bottom,carion2020detr} or image/video captioning~\cite{lu2019vilbert,sun2019videobert}.
% While very important,
We are interested in assessing how some of these advances can be combined for a holistic understanding of video clips.
% and how far can we reach with them.
% \mt{I've commented out the applications part below, can re-introduce if there's space}
% In particular, we foresee that successful systems could have wide-ranging applications from teaching embodied agents to understand and interact with the world~\cite{Young2020VisualIM},
% to video retrieval~\cite{Miech2019How2100M} and question-answering~\cite{howtovqa69m},
% and even fine-grained movie understanding~\cite{huang2020movie,moviegraphs}.

\begin{wrapfigure}{r}{0.6\linewidth}
\vspace{-0.4cm}
\centering
\includegraphics[width=1\linewidth]{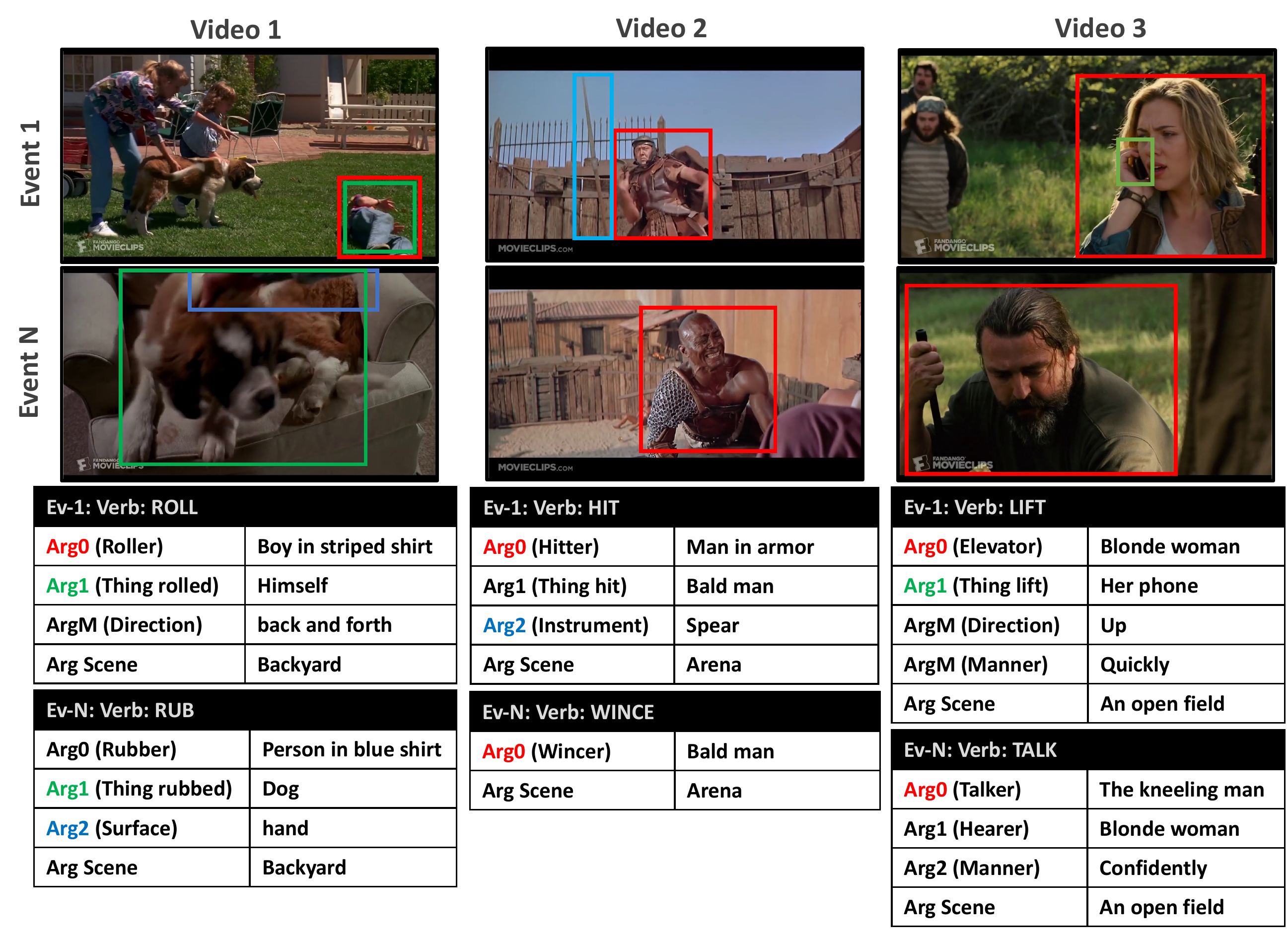}
\vspace{-0.6cm}
\caption{\textbf{Overview of GVSR:}
% Holistic understanding of video requires to localise and recognise all the salient events, and answer questions about it, like who is the agent, who is the patient, what is the location~\etc.GVSR affords this by recognising the actions, their corresponding roles, and localising them in the spatio-temporal domain. This is a challenging task as it requires to disambiguate between several roles that the same entity may take throughout the video, \eg~in Video 2 the \textit{bald man} is a \emph{patient} in event 1, but an \emph{agent} in event N. Colored arguments are highlighted in the image with boxes (figure best seen in color).
Given a video consisting of multiple events, GVSR requires recognising the action verbs, their corresponding roles, and localising them in the spatio-temporal domain.
This is a challenging task as it requires to disambiguate between several roles that the same entity may take in different events, \eg~in Video 2 the \textit{bald man} is a \emph{patient} in event 1, but an \emph{agent} in event N.
Moreover, the entities present in multiple events are co-referenced in all such events.
Colored arguments are grounded in the image with bounding boxes (figure best seen in colour).}
\label{fig:Teaser}
\vspace{-0.2cm}
\end{wrapfigure}

A recent and audacious step towards this goal is the work by Sadhu~\etal~\cite{sadhu2021vidsitu}.
They propose Video Situation Recognition (VidSitu), a structured prediction task over five short clips consisting of three sub-problems:
(i) recognizing the salient actions in the short clips;
(ii) predicting roles and their entities that are part of this action; and
(iii) modelling simple event relations such as enable or cause.
% To take a step in this endeavor, authors of \citep{sadhu2021vidsitu} proposed VidSitu, it involves multiple tasks to aid situation recognition in videos. 1) Recognizing and localizing multiple salient events in a video, i.e. action/verb classification. 2) Recognizing the entities taking part in it including actors, objects, locations with the roles they play. And, 3) Event relation.
Similar to the predecessor image situation recognition (imSitu~\cite{yatskar2016imsitu}), VidSitu is annotated using Semantic Role Labelling (SRL)~\cite{palmer2005propbank}.
A video (say 10s) is divided into multiple small events ($\sim$2s) and each event is associated with a salient action verb (\eg~\emph{hit}).
Each verb has a fixed set of roles or arguments, \eg~\emph{agent-Arg0}, \emph{patient-Arg1}, \emph{tool-Arg2}, \emph{location-ArgM(Location)}, \emph{manner-ArgM(manner)}, \etc, and each role is annotated with a free form text caption, \eg~\emph{agent: Blonde Woman}, as illustrated in Fig.~\ref{fig:Teaser}.

\mypara{Grounded VidSitu.}
VidSitu poses various challenges: long-tailed distribution of both verbs and text phrases, disambiguating the roles, overcoming semantic role-noun pair sparsity, and co-referencing of entities in the entire video.
Moreover, there is ambiguity in text phrases that refer to the same unique entity (\eg~``man in white shirt'' or ``man with brown hair''%
% or even simply ``man'' if there is no other male in the clip
).
% making the task very challenging.
% This ambiguity arises from the variability in the textual modality. The entity 'man' can be represented with multiple phrases, (eg: 'man in white shirt' or 'man with brown hair').
A model may fail to understand which attributes are important and may bias towards a specific caption (or pattern like shirt color), given the long-tailed distribution.
This is exacerbated when multiple entities (\eg~\emph{agent} and \emph{patient}) have similar attributes and the model predicts the same caption for them (see Fig.~\ref{fig:Teaser}).
To remove biases of the captioning module and gauge the model's ability to identify the role, we propose \emph{Grounded Video Situation Recognition} (GVSR) - an extension of the VidSitu task to include spatio-temporal grounding.
In addition to predicting the captions for the role-entity pairs, we now expect the structured output to contain spatio-temporal localization, currently posed as a weakly-supervised task.
% where we expect the 
% domain in a self supervised way.
% This eliminates the ambiguity present in the textual format. 
% \mt{challenges of weak grounding - removing this most likely}
% Self-supervised grounding poses several challenges.
% It requires to disentangle and model all the roles in a video, and ground every corresponding visual entity based on the supervision from text phrases only. This calls for global video level understanding with local disentangled role representations, and strong multimodal connections. 

\mypara{Joint structured prediction.}
Previous works~\cite{sadhu2021vidsitu, xiao2022hierarchical} modeled the VidSitu tasks separately, \eg~the ground-truth verb is fed to the SRL task.
% And require ground truths of previous tasks to model the next. 
This setup does \emph{not} allow for situation recognition on a new video clip without manual intervention.
%(providing the correct action label).
Instead, in this work, we focus on solving three tasks jointly:
(i) verb classification; (ii) SRL; and (iii) Grounding for SRL.
We ignore the original event relation prediction task in this work, as this can be performed later in a decoupled manner similar to~\cite{sadhu2021vidsitu}.

% and model the 3 tasks simultaneously. We propose role prediction followed by learnable role embeddings contextualised by event embeddings, to circumvent the requirement of ground truth verbs to model SRL. It enables end-to-end modelling and we call this new setup Grounded Situation Recognition in videos (GSRLV)

We propose \emph{\modelname{}}, a new three-stage transformer architecture that enables video understanding at a global level through self-attention across all video clips, and generates predictions for the above three tasks at an event level through localised event-role representations.
In the first stage, we use a Transformer encoder to align and contextualise 2D object features in addition to event-level video features.
These rich features are essential for grounded situation recognition, and are used to predict both the verb-role pairs %(using video features)
and entities. % (using object dependent).
In the second stage, a Transformer decoder models the role as a query, and applies cross-attention to find the best elements from the contextualised object features, also enabling visual grounding.
Finally, in stage three, we generate the captions for each role entity.
The three-stage network disentangles the three tasks and allows for end-to-end training.
% Additionally, cross-attention between role and object pairs naturally gives rise to visual grounding information. 

\mypara{Contributions summary.}
(i)~We present a new framework that combines grounding with SRL for end-to-end Grounded Video Situation Recognition (GVSR).
We will release the grounding annotations and also include them in the evaluation benchmark.
% and provide baselines for future work.
(ii)~We design a new three-stage transformer architecture for joint verb prediction, semantic-role labelling through caption generation, and weakly-supervised grounding of visual entities. 
(iii) We propose role prediction and use role queries contextualised by video embeddings for SRL, circumventing the requirement of ground-truth verbs or roles, enabling end-to-end GVSR.
(iv) At the encoder, we combine object features with video features and highlight multiple advantages enabling weakly-supervised grounding and improving the quality of SRL captions leading to a 22 points jump in CIDEr score in comparison to a video-only baseline~\cite{sadhu2021vidsitu}.
(v) Finally, we present extensive ablation experiments to analyze our model.
Our model achieves the state-of-the-art results on the VidSitu benchmark.

%% file: sections/2_related.tex
\section{Related Work}
\mypara{Image Situation Recognition.}
Situation Recognition in images was first proposed by~\cite{gupta2015visualSRL} where they created datasets to understand actions along with localisation of objects and people.
Another line of work, imSitu~\cite{yatskar2016imsitu} proposed situation recognition via semantic role labelling by leveraging linguistic frameworks, FrameNet~\cite{baker1998framenet} and WordNet~\cite{Miller1995wordnet} to formalize situations in the form of verb-role-noun triplets.
% detailed information about
Recently, grounding has been incorporated with image situation recognition~\cite{pratt2020gsr} to add a level of understanding for the predicted SRL.
Situation recognition requires global understanding of the entire scene, where the verbs, roles and nouns interact with each other to predict a coherent output.
Therefore several approaches used CRF~\cite{yatskar2016imsitu}, LSTMs~\cite{pratt2020gsr} and Graph neural networks~\cite{Li2017graphimsitu} to model the global dependencies among verb and roles.
Recently various Transformer~\cite{vaswani2017transformer} based methods have been proposed that claim large performance improvements~\cite{cho2022CoFormer,cho2021grimsituBMVC,wei2021grimsituAAAI}.

\mypara{Video Situation Recognition.}
Recently, imSitu was extended to videos as VidSitu~\cite{sadhu2021vidsitu}, a large scale video dataset based on short movie clips spanning multiple events.
% \del{They used Propbank~\cite{palmer2005propbank} to annotate the videos with event level verb-role pairs and free form text captions referring to each role.} 
Compared to image situation recognition, VidSRL not only requires understanding the action and the entities involved in a single frame, but also needs to coherently understand the entire video while predicting event-level verb-SRLs and co-referencing the entities participating across events.
Sadhu~\etal~\cite{sadhu2021vidsitu} propose to use standard video backbones for feature extraction followed by multiple but separate Transformers to model all the tasks individually, using ground-truth of previous the task to model the next.
A concurrent work to this submission, \cite{xiao2022hierarchical}, proposes to improve upon the video features by pretraining the low-level video backbone using contrastive learning objectives, and pretrain the high-level video contextualiser using event mask prediction tasks resulting in large performance improvements on SRL.
Our goals are different from the above two works, we propose to learn and predict all three tasks simultaneously.
% : verb and role classification, semantic role labelling, and grounding of roles simultaneously in an end-to-end fashion.
% In this work, we propose a new task Grounded SRL. 
To achieve this, we predict verb-role pairs on the fly and design a new role query contextualised by video embeddings to model SRL.
This eliminates the need for ground-truth verbs and enables end-to-end situation recognition in videos.
We also propose to learn contextualised object and video features enabling weakly-supervised grounding for SRL, which was not supported by previous works.

\mypara{Video Understanding.}
Video understanding is a broad area of research, dominantly involving tasks like
action recognition~\cite{carreira2017I3d, feichtenhofer2019slowfast, girdhar2019videoactiontrans, sun2018ActorRelNet, wang2016temporal, wu2019long}, 
localisation~\cite{lin2019bmn, lin2018bsn},
object grounding~\cite{sadhu2020video, yang2022tubedetr},
question answering~\cite{tapaswi2016movieqa, Yu2019ActivityVQA},
video captioning~\cite{rohrbach2017lsmdc},
and spatio-temporal detection~\cite{girdhar2019videoactiontrans, tapaswi2021ava}.
These tasks involve visual temporal understanding in a sparse uni-dimensional way.
% Where as, dense video understanding tasks like captioning \cite{zhou2018vidCAP} provides fine-grained details but in an unstructured way, which makes them difficult to evaluate and take decision upon.
In contrast, GVSR involves a hierarchy of tasks, coming together to provide a fixed structure, enabling dense situation recognition.
The proposed task requires global video understanding through event level predictions and fine-grained details to recognise all the entities involved, the roles they play, and simultaneously ground them.
Note that our work on grounding is different from classical spatio-temporal video grounding~\cite{zhang2020whereisit,yang2022tubedetr} or referring expressions based segmentation~\cite{khoreva2018vosre} as they require a text query as input.
In our case, both the text and the bounding box (grounding) are predicted jointly by the model.

% Moreover the predictions are co-referenced across the entire video to maintain coherency. 

%% file: sections/3_method.tex
\section{\modelname{} for Grounded Video Situation Recognition}

\begin{figure*}
\centering
\includegraphics[width=0.90\textwidth]{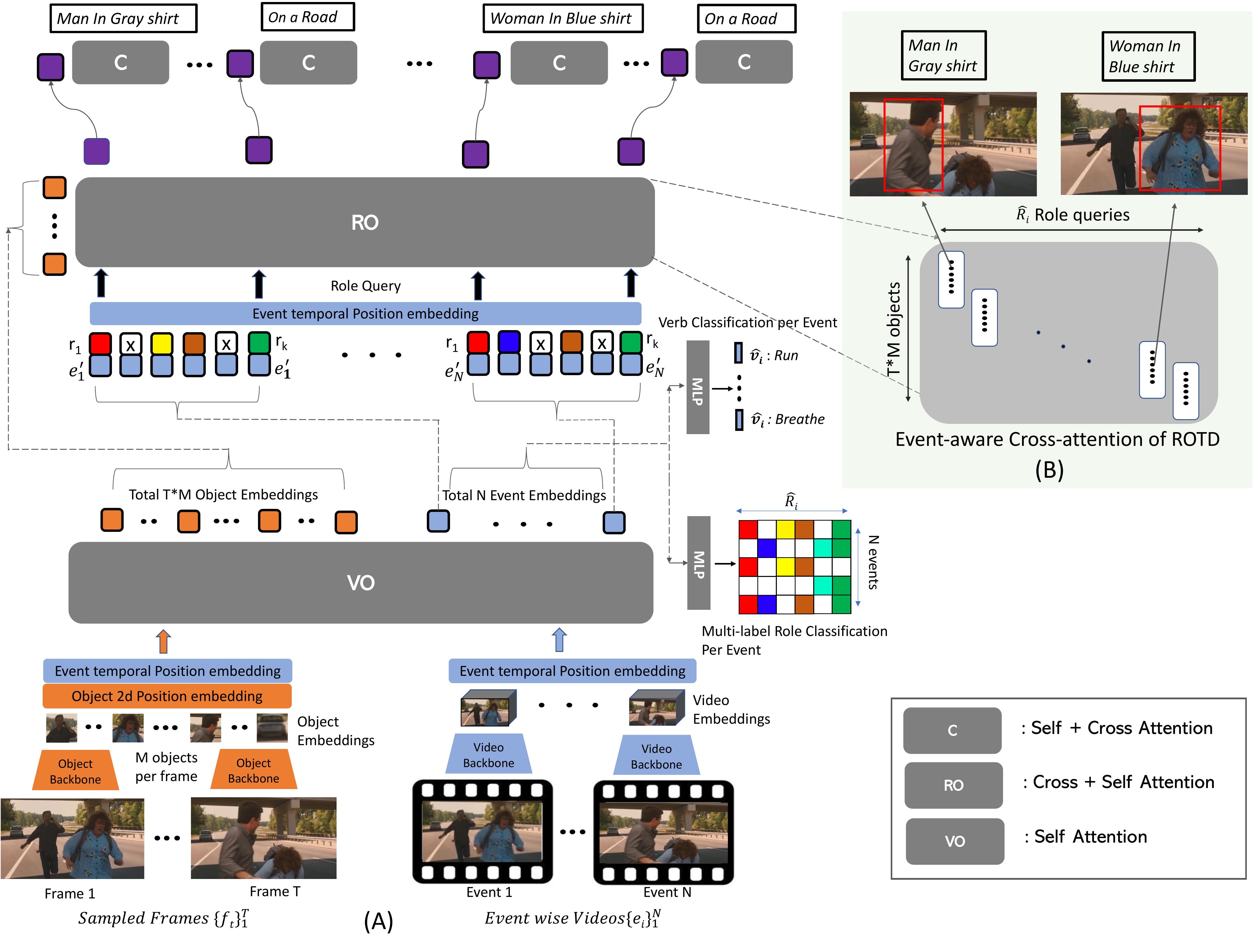}
\vspace{-0.3cm}
\caption{\textbf{VideoWhisperer}: We present a new 3-stage Transformer for GVSR.
Stage-1 learns the contextualised object and event embeddings through a video-object Transformer encoder (VO), that is used to predict the verb-role pairs for each event.
Stage-2 models all the predicted roles by creating role queries contextualised by event embeddings, and attends to all the object proposals through a role-object Transformer decoder (RO) to find the best entity that represents a role.
The output embeddings are fed to captioning Transformer decoder (C) to generate captions for each role.
Transformer RO's cross-attention ranks all the object proposals enabling localization for each role.}
\label{fig:Main_Model}
\vspace{-3mm}
\end{figure*}

We now present the details of our three stage Transformer model,~\emph{\modelname{}}.
A visual overview is presented in Fig.~\ref{fig:Main_Model}.
For brevity, we request the reader to refer to~\cite{vaswani2017transformer} for now popular details of self- and cross-attention layers used in Transformer encoders and decoders.

\mypara{Preliminaries.}
Given a video $V$ consisting of several short events $\mcE = \{e_i\}$, the complete situation in $V$, is characterised by 3 tasks.
(i)~Verb classification, requires predicting the action label $v_i$ associated with each event $e_i$;
(ii)~Semantic role labelling (SRL), involves guessing the nouns (captions) $\mcC_i = \{C_{ik}\}$ for various roles $\mcR_i = \{r | r \in \mcP(v_i) \forall r \in \mcR \}$ associated with the verb $v_i$.
$\mcP$ is a mapping function from verbs to a set of roles based on VidSitu (extended PropBank~\cite{palmer2005propbank}) and $\mcR$ is the set of all roles); and
(iii)~Spatio-temporal Grounding of each visual role-noun prediction $C_{ij}$ is formulated as selecting one among several bounding box proposals $\mcB$ obtained from sub-sampled keyframes of the video.
We evaluate this against ground-truth annotations done at a keyframe level.

% the form of caption generation; and
% (iii) Grounding nouns corresponding to each visual role.
% Each event $e_i$ is associated with one verb $v_i$, and the task is to predict all the verbs $V = \{v_i\}_i^N$ in the entire video. Each verb is associated with a fixed set of multiple roles. Let $p_i$ be the number of roles for each verb $v_i$, and  $R = \{r_{ji}\}_{j,i}^{p_i,N}$ be all the valid roles spanning all the events in a video $v$. The SRL task is to generate captions $ C = \{c_{ji}\}_{j,i}^{p_i,N}$ for the entities corresponding to all the roles $R$ in a video. For grounding SRL, the task is to ground all the entities corresponding to each valid visual role. 
% Where, boxes $B = \{B_1_1, ...,  {B}_{KN}_N\}$. 
% Previous works~\cite{sadhu2021vidsitu,xiao2022hierarchical} are limited to Verb classification and SRL only, they model both the tasks separately, and require ground truth verbs to model SRL. In contrast we model all the three tasks in and end to end setup without requiring ground truths of the previous task to model the next, and 

\subsection{Contextualised Video and Object Features (Stage 1)}

GVSR is a challenging task, that requires to coherently model spatio-temporal information to understand the salient action, determine the semantic role-noun pairs involved with the action, and simultaneously localise them.
Different from previous works that operate only on event level video features, we propose to model both the event and object level features simultaneously.
We use a pretrained video backbone $\phi_\text{vid}$ to extract event level video embeddings $\bx^e_i = \phi_\text{vid}( e_i )$.
% \begin{equation}
% \bx^e_i = \phi_\text{vid}( e_i ) \, .
% \end{equation}
For representing objects, we subsample frames $\mcF = \{f_t\}_{t=1}^T$ from the entire video $V$.
We use a pretrained object detector $\phi_\text{obj}$ and extract top $M$ object proposals from every frame.
The box locations (along with timestamp) and corresponding features are
\begin{equation}
\mcB = \{ b_{mt} \}, m = [1, \ldots, M], t = [1, \ldots, T] \,,
\quad \text{and} \quad
\{\bx^o_{mt}\}_{m=1}^M = \phi_\text{obj} ( f_t )
% \bx^o_{mt} = \phi_\text{obj} ( f_t, b_{mt} )
\quad \text{respectively}.
\end{equation}
The subset of frames associated with an event $e_i$ are computed based on the event's timestamps,
\begin{equation}
\mcF_i = \{ f_t | e_i^\text{start} \leq t \leq e_i^\text{end} \} \, .
\end{equation}
Specifically, at a sampling rate of 1fps, video $V$ of 10s, and events $e_i$ of 2s each, we associate 3 frames with each event such that the border frames are shared.
We can extend this association to all object proposals based on the frame in which they appear and denote this as $\mcB_i$.

% \begin{equation}
% F_1 =  \{f_1, f_2, f_3\} \in e_1; \: F_2 = \{f_3, f_4, f_5\} \in e_2; \: ... \:;\: F_N = \{f_{n-2}, f_{n-1}, t_{n}\} \in e_N
% \end{equation}
% Similarly Objects are also divided event wise, and the objects belonging to boundary frames are shared between the 2 neighbouring events. 
% \begin{equation}
% O_1 = \{o_{1,1}, ..., o_{3M,1}\} \:... \: O_N = \{o_{1,N}, ..., o_{3M,N}\}
% \end{equation}
% where $o_{j,i}$ represents object j extracted from the frames corresponding to event i: $F_i$

\mypara{Video-Object Transformer Encoder (VO).}
Since the object and video embeddings come from different spaces, we align and contextualise them with a Transformer encoder~\cite{vaswani2017transformer}.
Event-level position embeddings $\PE_i$ are added to both representations, event $\bx^e_i$ and object $\bx^o_{mt}$ ($t \in \mcF_i$).
In addition, 2D object position embeddings $\PE_{mt}$ are added to object embeddings $\bx^o_{mt}$.
Together, they help capture spatio-temporal information.
% We add learnable event-level temporal positional embeddings to both object and event embeddings. And 2D object positional embedding to object embeddings only. These positional embeddings provide a sense of temporal and spatial locations to the event and object representations.
The object and video tokens are passed through multiple self-attention layers to produce contextualised event and object embeddings:
\begin{equation}
[\ldots, \bo'_{mt}, \ldots, \be'_i, \ldots] = 
\text{Transformer}_\text{VO} \left( [\ldots, \bx_{mt}^o + \PE_i + \PE_{mt}, \ldots, \bx_i^e + \PE_i, \ldots ] \right) \, .
\end{equation}
%   \{e'_i\}_i^N, \{o'_{mt}\}_{m,t}^{M,T} = \text{Tx-VOTE}(\{emb_{{O}_{mt}}\}_{m,t}^{M,T}, \{emb_{{e}_i}\}_i^N) \, .

\mypara{Verb and role classification.}
Each contextualised event embedding $\be'_i$ is not only empowered to combine information across neighboring events but also focus on key objects that may be relevant.
We predict the action label for each event by passing them through a 1-hidden layer MLP,
% feed them through an MLP for They are fed to an MLP for verb classification. For each event $e_i$ we predict the corresponding verb $\hat{vb_i}$. 
\begin{equation}
\hat{v}_i = \text{MLP}_e( \be'_i ) \, .
\end{equation}

Each verb is associated with a fixed set of roles based on the mapping  $\mcP(\cdot)$.
This prior information is required to model the SRL task.
Previous works~\cite{sadhu2021vidsitu, xiao2022hierarchical} use ground-truth verbs to model SRL and predict both the roles and their corresponding entities.
While this setup allows for task specific modelling, it is not practical in the context of end-to-end video situation recognition.
To enable GVSR, we predict the relevant roles for each event circumventing the need for ground-truth verbs and mapped roles.
Again, we exploit the contextualised event embeddings and pass them through a role-prediction MLP and perform \emph{multi-label} role classification.
Essentially, we estimate the roles associated with an event as
\begin{equation}
\hat{\mcR}_i = \{ r | \sigma( \text{MLP}_r (\be'_i) ) > \theta_\text{role} \} \, ,
\end{equation}
% \{\hat{r_{ji}}\}_{j,i}^{p_i,N} = Sigmoid(MLP_2(\{e'_i\}_i^N))
where $\sigma(\cdot)$ is the sigmoid function and $\theta_\text{role}$ is a common threshold across all roles (typically set to 0.5).
Armed with verb and role predictions, $\hat{v}_i$ and $\hat{\mcR}_i$, we now look at localising the role-noun pairs and generating the SRL captions.

\subsection{Semantic Role Labelling with Grounding (Stage 2, 3)}

A major challenge in SRL is to disambiguate roles, as the same object (person) may take on different roles in the longer video $V$.
For example, if two people are conversing, the \emph{agent} and \emph{patient} roles will switch between \emph{speaker} and \emph{listener} over the course of the video.
Another challenge is to generate descriptive and distinctive captions for each role such that they refer to a specific entity.
We propose to use learnable role embeddings $\{\br_{ik}\}_{k=1}^{|\mcR_i|}$ which are capable of learning distinctive role representations.
As mentioned earlier, roles such as \emph{agent}, \emph{patient}, \emph{tool}, \emph{location}, \emph{manner}, \etc. ask further questions about the salient action.

\mypara{Creating role queries.}
Each role gets updated by the verb.
For example, for an action \emph{jump}, the \emph{agent} would be referred to as the \emph{jumper}.
We strengthen the role embeddings by adding the contextualised event embeddings to each role, instead of encoding ground-truth verbs.
This eliminates the dependency on the ground-truth verb-role pairs, and enables end-to-end GVSR.
Similar to the first stage (VO), we also add event-level temporal positional embeddings to obtain role \emph{query} vectors
% \begin{equation}
%     \{q_{ji}\}_{j,i}^{p_i,N} = \{\hat{r_{ji}}\}_{j,i}^{p_i,N} + \{\{{c\_{Ev}}_i\}_i^{N_j}\}_j^{pi} + Pos\_emb
% \end{equation}
\begin{equation}
\bq_{ik} = \br_{k} + \be'_i + \PE_i \, .
\end{equation}
Depending on the setting, $k$ can span all roles $\mcR$, ground-truth roles $\mcR_i$ or predicted roles $\hat{\mcR}_i$.

\mypara{Role-Object Transformer Decoder (RO).}
It is hard to achieve rich captions while using features learned for action recognition.
Different from prior works~\cite{sadhu2021vidsitu, xiao2022hierarchical}, we use fine-grained object level representations instead of relying on event-based video features.
We now describe the stage two of our \modelname{} model, the Transformer decoder for SRL.

% To facilitate descriptive captioning for the entities corresponding to semantic roles, it would require fine-grained object level representations.
% Instead, we rely on contextualised object features that are cross-attention networks to identify

Our Transformer decoder uses semantic roles as queries and object proposal representations as keys and values.
Through the cross-attention layer, the event-aware role query attends to contextualised object embeddings and finds the best objects that represent each role.
% through event-aware cross-attention, and find the best objects that represent each role.
We incorporate an event-based attention mask, that limits the roles corresponding to an event to search for objects localised in the same event, while masking out objects from other events.
Cross-attention captures local event-level role-object interactions while the self-attention captures the global video level understanding allowing event roles to share information with each other.
% and to maintain coherence among the roles.

We formulate event-aware cross-attention as follows.
We first define the query, key, and value tokens fed to the cross-attention layer as
\begin{equation}
\bq'_{ik} = W_Q \bq_{ik}, \quad
\bk'_{mt} = W_K \bo'_{mt}, \quad \text{and} \quad
\bv'_{mt} = W_V \bo'_{mt} \, .
\end{equation}
Here, $W_{[Q|K|V]}$ are learnable linear layers.
% Now we formulate the event-aware cross attention: We define $Q = Lin_Q(\{q_{ji}\}_{j,i}^{p_i,N})$, $K = Lin_K(\{o'_{mt}\}_{m,t}^{M,T})$ and $V = Lin_V(\{o'_{mt}\}_{m,t}^{M,T})$. Where $Lin_Q$, $Lin_K$ and, $Lin_V$, represents the linear mappings from input to query, key and, value embeddings, i.e. $Q, K, and, V$ respectively. 
Next, we apply a mask while computing cross-attention to obtain contextualised role embeddings as
% \begin{equation}
%     Mask(i,j)= 
% {\begin{cases}
%     1,& \text{if } Obj\ j \in F_i\\
%     0,              & \text{otherwise}
% \end{cases}}    
% \end{equation}
\begin{equation}
\br'_{ik} = \sum_{mt} \alpha_{mt} \bv'_{mt}, \quad \text{where} \quad
\alpha_{mt} = \text{softmax}_{mt} ( \langle \bq'_{ik}, \bk'_{mt} \rangle \cdot \mathds{1}(f_t \in \mcF_i) ) \, ,
\end{equation}
where $\langle \cdot, \cdot \rangle$ is an inner product and $\mathds{1}(\cdot)$ is an indicator function with value 1 when true and $-\infty$ otherwise to ensure that the cross-attention is applied only to the boxes $\mcB_i$, whose frames $f_t$ appear within the same event $e_i$.

% $Mask \in \{0,1\}^{Q_l\times K_l}$, where, $Q_l = Len(Q) = \sum_{i=1}^{N} p_i $ i.e. all the valid roles. and $K_l = len(K) = T*M $ i.e. all the object proposals. 
After multiple layers of cross- and self-attention, the role query extracts objects that best represent the entities for each role.
% And we get the transformed semantic role embeddings (SRE).
\begin{equation}
[\ldots, \bz_{ik}, \ldots] = \text{Transformer}_\text{RO} ([\ldots, \bq_{ik}, \ldots; \ldots, \bo'_{mt}, \ldots]) \, .
\end{equation}

\mypara{Captioning Transformer Decoder (C).}
The final stage of our model is a caption generation module.
Specifically, we use another Transformer decoder~\cite{vaswani2017transformer} whose input context is the output role embedding $\bz_{ik}$ from the previous stage and unroll predictions in an autoregressive manner.
% \del{Different from prior work, captions are generated separately for each role in parallel, however, all the necessary information has already been shared through multiple layers of the second stage $\text{Transformer}_\text{ROTD}$.}
\begin{equation}
\hat{C}_{ik} = \text{Transformer}_\text{C} ( \bz_{ik} ) \, .
\end{equation}
The role-object decoder in stage 2 shares all the necessary information through self-attention, and allows us to generate the captions for all the roles in parallel; while~\cite{sadhu2021vidsitu, xiao2022hierarchical} generate captions sequentially
, \ie~for a given event, the caption for role $k$ is decoded only after the caption for role $k-1$.
This makes \modelname{} efficient with a wall-clock runtime of 0.4s for inference on a 10s video, while the baseline~\cite{sadhu2021vidsitu} requires 0.94 seconds.

\mypara{Grounded Semantic Role Labelling.}
The entire model is designed in a way to naturally provide SRL with grounding in a weakly-supervised way, without the need for ground-truth bounding boxes during training.
% As mentioned above, we incorporate event-aware cross-attention for the Role query to attend to objects associated with the corresponding event.
Cross-attention through the Transformer decoder RO scores and ranks all the objects based on the role-object relevance at every layer.
We extract the cross-attention scores $\alpha_{mt}$ for each role $k$ and event $e_i$ from the final layer of $\text{Transformer}_\text{RO}$, and identify the highest scoring box and the corresponding timestep as
% $\hat{b}^* = \arg\max_{b_{mt}} \alpha_{mt}$.
\begin{equation}
\label{eq:boxargmax}
\hat{b}_m^*, \hat{b}_t^* = \arg\max_{m, t} \alpha_{mt} \, .
\end{equation}

\subsection{Training and Inference}
% We briefly discuss how we can train or obtain predictions from our model.

\mypara{Training.}
\modelname{} can be trained in an end-to-end fashion, with three losses.
The first two losses, CrossEntropy and BinaryCrossEntropy, are tapped from the contextualis    ed event embeddings and primarily impact the Video-Object Transformer encoder
\begin{equation}
L^\text{verb}_i = CE(\hat{v}_i, v_i) \quad \text{and} \quad
L^\text{role}_i = \sum_{r \in \mcR_i} BCE(r \in \hat{\mcR}_i, r \in \mcR_i) \, .
\end{equation}
The final component is derived from the ground-truth captions and helps produce meaningful SRL outputs.
This is also the source of weak supervision for the grounding task,
\begin{equation}
L^\text{caption}_{ik} = \sum_w CE(\hat{C}^w_{ik} , C^w_{ik}) \, ,
\end{equation}
where the loss is applied in an autoregressive manner to each predicted word $w$.
The combined loss for any training video $V$ is given by
\begin{equation}
\mcL = \sum_i L^\text{verb}_i + \sum_i L^\text{role}_i + \sum_{ik} L^\text{caption}_{ik} \, .
\end{equation}

% $L_{verb}$, 2) Multilabel Role classification loss: $L_{role}$, 3) SRL Captioning loss: $L_{caption}$.

% $L_{verb} = \dfrac{1}{N} \sum_{i=1}^{N} CE(\hat{v_i} , \: v_i); \: \: \:  L_{role} = \dfrac{1}{N} \dfrac{1}{p_i}  \sum_{i=1}^{N} \sum_{j=1}^{p_i} BCE(\hat{r_{ji}} ,\:
% r_{ji}) \: ;$ 

% $L_{caption} = \dfrac{1}{N} \dfrac{1}{p_i}  \sum_{i=1}^{N} \sum_{j=1}^{p_i} CE(\hat{c_{ji}} ,\: c_{ji}).$ Final loss term is $L_{verb} + L_{role} + L_{caption}$

\mypara{Inference.}
At test time, we split the video $V$ into similar events $e_i$ and predict verbs $\hat{v}_i$ and roles $\hat{\mcR}_i$ for the same.
Here, we have two options:
(i)~we can use the predicted verb and obtain the corresponding roles using a ground-truth mapping between verbs and roles $\mcP(\hat{v}_i)$, or
(ii)~only predict captions for the predicted roles $\hat{\mcR}_i$.
We show the impact of these design choices through experiments.

%% file: sections/4_experiments.tex
\section{Experiments}
We evaluate our model in two main settings.
(i) This setup mimics VidSitu~\cite{sadhu2021vidsitu}, where tasks are evaluated separately. We primarily focus on (a)~Verb prediction, (b)~SRL and (c)~Grounded SRL.
This setting uses ground-truth verb-role pairs for modelling (b) and (c).
(ii) End-to-end GVSR, where all the three tasks are modelled together without using ground truth verb-roles.

\mypara{Dataset.}
We evaluate our model on the VidSitu~\cite{sadhu2021vidsitu} dataset that consists of 29k videos (23.6k train, 1.3k val, and others in task-specific test sets) collected from a diverse set of 3k movies.
% The train and validation sets consists of 23,626 and 1,326 samples respectively, while the remainder are reserved as task-specific test sets.
All videos are truncated to 10 seconds, have 5 events spanning 2 seconds each and are tagged with verb and SRL annotations.
There are a total of 1560 verb classes and each verb is associated with a fixed set of roles among 11 possible options, however not all are used for evaluation due to noisy annotations (we follow the protocol by~\cite{sadhu2021vidsitu}).
For each role the corresponding value is a free-form caption.
% There are 11 roles: Arg-0, Arg-1, Arg-2, Arg-3, Arg-4, Arg-Manner, Arg-Location, Arg-Direction, Arg-Purpose, Arg-Goal, Arg-Scene.
% Removed Arg-ADV

\mypara{Metrics.}
For verb prediction, we report Acc@K, \ie~event accuracy considering 10 ground-truth verbs and top-K model predictions and Macro-Averaged Verb Recall@K.
For SRL we report CIDEr~\cite{Cider},
CIDEr-Vb: Macro-averaged across verbs,
CIDEr-Arg: Macro-averaged across roles,
LEA~\cite{moosavi2016-LEA}, and
ROUGE-L~\cite{lin2004rouge}.
For more details on the metrics pleas refer to~\cite{sadhu2021vidsitu}.

\mypara{Implementation details.}
We implement our model in Pytorch~\cite{paszke2019pytorch}.
% Each video in VidSitu~\cite{sadhu2021vidsitu} dataset spans 5 events.
We extract event (video) features from a pretrained SlowFast model~\cite{feichtenhofer2019slowfast} for video representation (provided by~\cite{sadhu2021vidsitu}).
For object features, we use a FasterRCNN model~\cite{ren2015faster} provided by BUTD~\cite{anderson2018bottom} pretrained on the Visual Genome dataset~\cite{krishna2017visual}.
We sample frames at 1 fps from a 10 second video, resulting in $T = 11$ frames.
We extract top $M = 15$ boxes from each frame, resulting in 165 objects per video.

All the three Transformers have the same configurations - they have 3 layers with 8 attention heads, and hidden dimension 1024.
We use the tokenizer and vocabulary provided by VidSitu~\cite{sadhu2021vidsitu} which uses byte pair encoding.
We have 3 types of learnable embeddings:
(i) event position embeddings $\PE_i$ with 5 positions corresponding to each event in time;
(ii) object localization 2D spatial embedding; and
(iii) role embeddings, for each of the 11 roles.
The verb classification MLP has a single hidden layer of 2048 d and produces an output across all 1560 verbs.
The role classification MLP also has a single hidden layer of 1024 d and produces output in a multi-label setup for all the 11 roles mentioned above.
We threshold role prediction scores with $\theta_\text{role} = 0.5$.

We use the Adam optimizer~\cite{kingma2014adam} with a learning rate of $10^{-4}$ to train the whole model end-to-end.
As we use pretrained features, we train our model on a single RTX-2080 GPU, batch size of 16.

\subsection{Grounding SRL: Annotation and Evaluation}
As free form captions and their evaluation can be ambiguous, we propose
% an SRL model that generates same captions for similar looking entities, even when it is able to correctly disambiguate them. In order to overcome this ambiguity we propose
to simultaneously ground each correct role in the spatio-temporal domain.
To evaluate grounding performance, we obtain annotations on the validation set.
We select the same $T = 11$ frames that are fed to our model sampled at 1fps.
For each frame, we ask annotators to see if the visual roles (\emph{agent}, \emph{patient}, \emph{instrument}), can be identified by drawing a bounding box around them using the CVAT tool~\cite{cvat} (see Appendix~\ref{supp:sec:annotations} for a thorough discussion).
For each event $i$ and role $k$, we consider all valid boxes and create a dictionary of annotations $\mcG_{ik}$ with keys as frame number and value as bounding box.
% accumulates all frames and $\mcG^\text{box}_{ik}(t)$ returns bounding box coordinates for frame $t$.
% We map all the visual roles with the corresponding frames and bounding boxes. Let there be F frames where the role exists with B bounding boxes.
% Ground truth frame-box set for a role $r$ in a video $v$ is given by: $G^{rv} = \{(frame_f, box_b)\}_{f,b}^{F,B} \:\: \forall \: r \in R^v_G, v \in V)$, Where $R^v_G = \{ roles \: with \: grounding \: in \: video \: v\}$.
During prediction, for each role $r \in \hat{\mcR}_i$, we extract the highest scoring bounding box as in Eq.~\ref{eq:boxargmax}.
The Intersection-over-Union (IoU) metric for an event consists of two terms.
The first checks if the selected frame appears in the ground-truth dictionary, while the second compares if the predicted box has an overlap greater than $\theta$ with the ground-truth annotation,
% box: $P^{rv}_b$ and the frame: $P^{rv}_f$ it belongs to. For each visual role we check if the predicted frame is in the set of ground truth frames, then calculate IoU of the predicted and ground-truth box.
\begin{equation}
\text{IoU} @ \theta = \frac{1}{|\mcR_i|} \sum_{k=1}^{|\mcR_i|} \mathds{1} [\hat{b}^*_t \in \mcG_{ik}] \cdot \mathds{1} [\text{IoU}(\hat{b}^*_m, \mcG_{ik}[t]) > \theta] \, .
\end{equation}
% \begin{equation}
%   IoU@\theta = \frac{1}{|V|} \sum_{v} \dfrac{1}{|R^v_G|} \sum_{r \in R_G^v} \mathds{1} (P_f^{rv} \in G_f^{rv}) \cdot \mathds{1} [IoU(P_b^{rv}, G_{fb}^{rv})> \theta]
% \end{equation}

\begin{wraptable}{r}{0.71\linewidth}
% \begin{table}[h]
\tabcolsep=0.08cm
\vspace{-0.7cm}
\caption{Architecture ablations.
All the models use event-aware cross-attention.
+ indicates stages of the model.
V: Video encoder,
VO: Video-Object encoder,
VOR: Video-Object-Role encoder,
RV: Role-Video decoder,
RO: Role-Object decoder, and
C: Captioning Transformer.}
\label{tab:table1}
\begin{tabular}{lll ccc}
\toprule
\# & Architecture & Query Emb. & CIDEr & IoU@0.3 & IoU@0.5 \\
\midrule
1 & RV + C & Role + GT-verb & 47.91 \scriptsize{$\pm$ 0.53} & - & - \\
2 & RO + C & Role + GT-verb & {\bf 70.48} \scriptsize{$\pm$ 1.09} & 0.14 \scriptsize{$\pm$ 0.01} & 0.06 \scriptsize{$\pm$ 0.003} \\
3 & VOR + C & Role + Event & 67.4 \scriptsize{$\pm$ 0.81} & 0.22 \scriptsize{$\pm$ 0.00} & 0.09 \scriptsize{$\pm$ 0.002} \\
\midrule
4 & V + RO + C & Role + Event & 69.15 \scriptsize{$\pm$ 0.62} & 0.23 \scriptsize{$\pm$ 0.03} & 0.09 \scriptsize{$\pm$ 0.01} \\
5 & VO + RO + C & Role + Event & 68.54 \scriptsize{$\pm$ 0.48} & {\bf 0.29} \scriptsize{$\pm$ 0.013} & {\bf 0.12} \scriptsize{$\pm$ 0.01} \\
\bottomrule
\end{tabular}
\vspace{-0.3cm}
\end{wraptable}

\subsection{Grounded SRL Ablations}
We analyze the impact of architecture choices, role query embeddings, and applying a mask in the cross-attention of the role-object decoder.
All ablations in this section assume access to the ground-truth verb or roles as this allows us to analyze the effect of various design choices.
Similar to~\cite{xiao2022hierarchical} we observe large variance across runs, therefore we report the average accuracy and the standard deviation over 3 runs for all the ablation experiments and 10 runs for the proposed model (VO+RO+C).
% We will report performance over all metrics in the Appendix.
% by keeping the input verb-role pairs constant.
% This is comparable to~\cite{sadhu2021vidsitu} as 

\mypara{Architecture design.}
We present SRL and grounding results in Table~\ref{tab:table1}.
Rows 1 and 2 use a two-stage Transformer decoder (ignoring the bottom video-object encoder).
As there is no event embedding $\be'_i$, role queries are augmented with ground-truth verb embedding.
Using role-object pairs (RO) is critical for good performance on captioning as compared to role-video (RV), CIDEr 70.48 vs. 47.91.
Moreover, using objects enables weakly-supervised grounding.
Row 3 represents a simple Transformer encoder that uses self-attention to model all the video events, objects, and roles (VOR) jointly.
As before, role-object attention scores are used to predict grounding.
Incorporating videos and objects together improves the grounding performance. 

We switch from a two-stage to a three-stage model between rows 1, 2, 3 vs. 4 and 5.
Rows 2 vs. 5 illustrates the impact of including the video-object encoder.
We see a significant improvement in grounding performance 0.14 to 0.29 for IoU@0.3 and 0.06 to 0.12 for IoU@0.5 without significantly affecting captioning performance.
Similarly, rows 4 vs. 5 demonstrate the impact of contextualizing object embeddings by events.
In particular, using contextualised object representations $\bo'_{mt}$ seems to help as compared against base features $\bx^o_{mt}$.

% We ablate on various architectural choices, showing the importance of each component. We choose the role query: role + event embeddings and apply verb loss if the contextualised event features from the video encoder are available, else we use ground truth verb query to contextualise the role query with action information.
% As can be seen in Table \ref{tab:table1} there is a sharp jump in performance when using object features instead of video. This shows that fine-grained object features are necessary for generating correct descriptive captions. (RO+C) performs best on CIDEr, this little jump over (V+RO+C) is due to the fact, that the model only focuses on SRL task without the verb prediction, which slightly interferes with the SRL task. This shows that object features are good enough for greedily predicting descriptive captions. But in the absence of VO Transformer, the object embeddings and the role query in (RO+C) are not contextualised with event embeddings. Therefore, the objects loses association with fine-grained event information, which is necessary to ground the right roles in context of the action. Hence, (RO+C) perofrms poorly on GSRL. With video encoder (V+RO+C) we see an improvement in grounding accuracy, since the role query is now contextualised with event embeddings. In (VO+RO+C) both the object embeddings and role query are contextualised with event embeddings, giving the highest grounding accuracy.

\begin{wraptable}{r}{0.58\linewidth}
\tabcolsep=0.1cm
\vspace{-0.6cm}
\caption{Comparing role query embeddings.}
\label{tab:table2}
\begin{tabular}{llccc}
\toprule
\#  & Query Emb. & CIDEr & IoU@0.3 & IoU@0.5 \\
\midrule
1 & Role only & 68.61 \scriptsize{$\pm$ 0.61} & 0.27 \scriptsize{$\pm$ 0.011} & 0.11 \scriptsize{$\pm$ 0.009} \\
2 & Role + GT-verb & 68.71 \scriptsize{$\pm$ 1.06} & 0.25 \scriptsize{$\pm$ 0.02} & 0.10 \scriptsize{$\pm$ 0.01} \\
3 & Role + Event & 68.54 \scriptsize{$\pm$ 0.48} & {\bf 0.29} \scriptsize{$\pm$ 0.013} & {\bf 0.12} \scriptsize{$\pm$ 0.01} \\
\bottomrule
\end{tabular}
\vspace{-0.3cm}
\end{wraptable} 

\mypara{Role query embeddings design.}
Prior works in situation recognition~\cite{cho2021grimsituBMVC,sadhu2021vidsitu,wei2021grimsituAAAI} use verb embeddings to identify entities from both images or videos.
In this ablation, we show that instead of learning verb embeddings that only capture the uni-dimensional meaning of a verb and ignore the entities involved, event (or video) embeddings remember details and are suitable for SRL.
In fact, Table~\ref{tab:table2} (architecture: VO + RO + C) row 2 vs. 3 show that event embeddings are comparable and slightly better than GT-verb embeddings when evaluated on SRL and Grounding respectively, eliminating the need for GT verbs.
Somewhat surprisingly, we see that the role embeddings alone perform quite well.
We believe this may be due to role embeddings
(i)~capture the generic meaning like \emph{agent} and \emph{patient} and can generate the correct entities irrespective of the action information; and
(ii)~the role query attends to object features which are contextualised by video information, so the objects may carry some action information with them.

\begin{wraptable}{r}{0.44\linewidth}
\vspace{-0.6cm}
\tabcolsep=0.1cm
\caption{Impact of masking in RO decoder.}
\label{tab:table3}
\begin{tabular}{l ccc}
\toprule
Mask & CIDEr & IoU@0.3 & IoU@0.5 \\
\midrule
No & 67.02 \scriptsize{$\pm$ 0.51} & 0.25 \scriptsize{$\pm$ 0.02} & 0.10 \scriptsize{$\pm$ 0.012} \\ 
Yes & {\bf 68.54} \scriptsize{$\pm$ 0.48} & {\bf 0.29} \scriptsize{$\pm$ 0.013} & {\bf 0.12} \scriptsize{$\pm$ 0.01} \\ 
\bottomrule
\end{tabular}
\vspace{-0.2cm}
\end{wraptable} 

\mypara{Masked cross-Attention in RO decoder.}
We use masking in event-aware cross-attention to ensure that the roles of an event attend to objects coming from the same event.
As seen in Table~\ref{tab:table3} (model: VO + RO + C, query is role + event embedding), this reduces the object pool to search from and improves both the SRL and Grounding performance.

\subsection{SRL SoTA comparison}
In Table~\ref{tab:table4}, we compare our results against VidSitu~\cite{sadhu2021vidsitu} and a concurrent work that uses far better features~\cite{xiao2022hierarchical}.
We reproduce results for VidSitu~\cite{sadhu2021vidsitu} by teacher-forcing the ground-truth role pairs to make a fair comparison while results for work~\cite{xiao2022hierarchical} are as reported in their paper.
Nevertheless, we achieve state-of-the-art performance with a 22 points gain in CIDEr score over~\cite{sadhu2021vidsitu} and a 8 point gain over~\cite{xiao2022hierarchical}, while using features from~\cite{sadhu2021vidsitu}.
Moreover, our model allows grounding, something not afforded by the previous approaches.

\begin{table}[h!]
\begin{center}
\tabcolsep=0.10cm
\vspace{-0.2cm}
\caption{SoTA comparison, results for SRL and grounding with GT verb and role pairs.}
\label{tab:table4}
\vspace{-0.2cm}
\scalebox{0.85}{
\begin{tabular}{l c cc c c cc}
\toprule
Method & CIDEr & C-Vb & C-Arg & R-L & Lea & IoU@0.3 & IoU@0.5 \\
\midrule
SlowFast+TxE+TxD~\cite{sadhu2021vidsitu} & 46.01 & 56.37  & 43.58 & 43.04 & {\bf 50.89} & - & - \\
Slow-D+TxE+TxD~\cite{xiao2022hierarchical} & 60.34 \scriptsize{$\pm$ 0.75} & 69.12 \scriptsize{$\pm$ 1.43} & 53.87 \scriptsize{$\pm$ 0.97} & 43.77 \scriptsize{$\pm$ 0.38} & 46.77 \scriptsize{$\pm$ 0.61} & - & - \\
% \midrule
% RV+C & 48.03 & 56.92 & 45.51 & 41.98 & 55.60 & - & - \\
% RO+C & 69.08 & 77.89 & 64.73 & 45.59 & 48.30 & 0.21 & 0.08 \\
% V+RO+C & 68.08 & 77.90 & 60.00 & 45.20 & 46.11 & 0.23 & 0.11 \\
\modelname{} (Ours) & {\bf 68.54} \scriptsize{$\pm$ 0.48} & {\bf 77.48} \scriptsize{$\pm$ 1.52} & {\bf 61.55} \scriptsize{$\pm$ 0.79} & {\bf 45.70} \scriptsize{$\pm$ 0.30} & 47.54 \scriptsize{$\pm$ 0.55} & {\bf 0.29} \scriptsize{$\pm$ 0.013} & {\bf 0.12} \scriptsize{$\pm$ 0.01} \\
% \modelname{} (Ours) & {$\bm{68.54 \pm 0.48}$} & {$\bm{77.48 \pm 1.52}$} & {$\bm{61.55 \pm 0.79}$} & {$\bm{45.70 \pm 0.30}$} & $47.54 \pm 0.55$ & {$\bm{0.29 \pm 0.013}$} & {$\bm{0.12 \pm 0.01}$} \\
\midrule
\textcolor{gray}{Human Level} & \textcolor{gray}{84.85} & \textcolor{gray}{91.7} & \textcolor{gray}{80.15} & \textcolor{gray}{39.77} & \textcolor{gray}{70.33} & - & - \\ 
\bottomrule
\end{tabular}}
\end{center}
\vspace{-0.5cm}
\end{table}

\subsection{GVSR: Joint Prediction of Video Situations}
\label{subsec:exp:gvsr}
The primary goal of our work is to enable joint prediction of the verb, roles, entities, and grounding.
% In this section, we first discuss verb prediction performance.
% Then, we evaluate role prediction and it's impact on Grounded SRL, and end this section by presenting the first method and results for joint prediction of the complete GVSR pipeline.
% end-to-end GVSR, we need to predict verb-role pairs and
% prediction of verb-role pairs and their corresponding grounded entities.

\mypara{Verb prediction}
is an extremely challenging problem due to the long-tail nature of the dataset.
% In fact, to alleviate this challenge, verb metrics are computed by comparing predictions against 10 ground-truth human annotations.
In Table~\ref{tab:table6}, we evaluate verb prediction performance when training the model for verb prediction only (rows 1-3) or training it jointly for GVSR (rows 4, 5).
Using a simple video-only transformer encoder boosts performance over independent predictions for the five event clips (46.8\% to 48.8\%, rows 1 vs. 2).
Including objects through the video-object encoder (row 3) provides an additional boost resulting in the highest performance at 49.73\% on Accuracy@1.

\begin{wraptable}{r}{0.42\linewidth}
\vspace{-0.3cm}
\tabcolsep=0.07cm
\caption{Verb prediction performance.
Rows 1-3 train only for verb prediction.
Rows 4, 5 are trained for GVSR.}
\label{tab:table6}
\begin{tabular}{llccc}
\toprule
\# & Architecture & Acc@1 & Acc@5 & Rec@5 \\
\midrule
1 & Baseline~\cite{sadhu2021vidsitu} & 46.79 & 75.90 & 23.38 \\
2 & V & 48.82 & 78.01 & 23.32 \\
3 & VO & {\bf 49.73} & {\bf 78.72} & 24.72 \\
\midrule
4 & V + RV + C & 40.83 & 70.73 & 24.37 \\
5 & VO + RO + C & 45.06 & 75.59 & {\bf 25.25} \\ 
\bottomrule
\end{tabular}
\vspace{-0.3cm}
\end{wraptable}

A similar improvement is observed in rows 4 to 5 (V vs. VO stage 1 encoder).
Interestingly, the reduced performance of rows 4 and 5 as compared against rows 1-3 is primarily because the best epoch corresponding to the highest verb accuracy does not coincide with highest SRL performance.
Hence, while the verb Accuracy@1 of the GVSR model does reach 49\% during training it degrades subsequently due to overfitting.
Nevertheless, we observe that the macro-averaged Recall@5 is highest for our model, indicating that our model focuses on all verbs rather than just the dominant classes.
In Appendix~\ref{sec:supp:longtail-verbs}, we show the challenges of the large imbalance and perform experiments that indicate that classic re-weighting or re-sampling methods are unable to improve performance in a meaningful mannner.
Addressing this aspect is left for future work.

% from 46.79 \cite{sadhu2021vidsitu} to 49.73, when a Video Transformer encoder is used and each event attends to all the events through multiple self attentions. The accuracy further improves, when the full VO Transformer is used, where event features are contextualised with object features. In setting 2, Verb accuracy slightly drops when compared with \cite{sadhu2021vidsitu}. This is due to the fact that, joint training calls for joint optimization for both the tasks. Training SRL with Verb classification leads to overfitting of verbs, since during training we observe that verb accuracy reaches to 49 at early epochs and drops down to 45.06 till the SRL converges.

\mypara{Understanding role prediction.}
The verb-role prediction accuracy is crucial for GVSR, since the SRL task is modelled on role-queries.
In Table~\ref{tab:table5} we analyse role prediction in various settings to understand its effect on SRL.
Previous work~\cite{sadhu2021vidsitu} used ground-truth verbs for SRL, while roles and their entities or values are predicted sequentially.
This setting is termed ``GT, Pred'' (row 2) as it uses the ground-truth verb but predicts the roles.
We argue that as the verb-role mapping $\mcP$ is a deterministic lookup table, this setting is less interesting.
% We can provide both the ground truth verb and roles to model SRL,
We enforce a ``GT, GT'' setting with ground-truth verbs and roles in~\cite{sadhu2021vidsitu} by teacher-forcing the GT roles while unrolling role-noun predictions (row 1).
Another setting is where the verb is predicted and roles are obtained via lookup, ``Pred, GT map'' (row 3).
Note that this enables end-to-end SRL, albeit in two steps.
The last setting, ``Pred, Pred'' predicts both verb and role on-the-fly (row 4).

\begin{wraptable}{r}{0.60\linewidth}
\vspace{-0.6cm}
\tabcolsep=0.08cm
\caption{Role prediction in various settings. Role F1 is the F1 score averaged over all role classes.}
\label{tab:table5}
\begin{tabular}{llcc ccc}
\toprule
\# & Architecture & Verb & Role & V. Acc@1 & Role F1 & CIDEr \\
\midrule
1 & \multirow{4}{*}{VidSitu~\cite{sadhu2021vidsitu}} & GT & GT & - & - & 46.01 \\
2 & & GT & Pred & - & 0.88 & 45.52 \\
3 & & Pred & GT map & 46.79 & - & 29.93 \\
4 & & Pred & Pred & 46.79 & 0.47 & 30.33 \\
\midrule
5 & RO+C    & GT & GT & -  & - & 70.48 \\
6 & VO+RO+C & Pred &GT & 45.06 & - & 68.54 \\
7 & VO+RO+C & Pred & GT map & 45.06 & - & 51.24 \\
8 & VO+RO+C & Pred & Pred & 44.05 & 0.44 & 52.30 \\
\bottomrule
\end{tabular}
\vspace{-0.2cm}
\end{wraptable}

Comparing within variants of~\cite{sadhu2021vidsitu}, surprisingly, row 1 does not perform much better than row 2 on CIDEr.
This may be because the model is trained on GT verbs and is able to predict most of the roles correctly (row 2, Role F1 = 0.88).
Subsequently, both rows 3 and 4 show a large performance reduction indicating the over-reliance on ground-truth verb.
We see similar trends for our models.
Rows 7 and 8 with predicted verb-role pairs lead to reduced SRL performance as compared against rows 5 and 6.
Nevertheless, our ``Pred, Pred'' CIDEr score of 52.3 is still higher than the baseline ``GT, GT'' at 46.0.
Appendix~\ref{supp:subsec:role_prediction} discusses further challenges of multi-label and imbalance in predicting roles.

\begin{wraptable}{r}{0.62\linewidth}
\vspace{-0.6cm}
\tabcolsep=0.08cm
\caption{GVSR: Results for end-to-end situation recognition.
Our model architecture is VO + RO + C.}
\label{tab:table7}
% \resizebox{\textwidth}{!}{
\begin{tabular}{l ccc c c cc}
\toprule
\multirow{2}{*}{Model} & \multicolumn{3}{c}{Prediction} & Verb & \multirow{2}{*}{CIDEr} & \multicolumn{2}{c}{IoU} \\
 & Verb & Role & SRL & Acc@1 & & 0.3 & 0.5\\
\midrule
VidSitu~\cite{sadhu2021vidsitu} & \checkmark & \checkmark & \checkmark & 46.79 & 30.33 & - & - \\
\midrule
\multirow{2}{*}{\modelname{}} & \checkmark & \checkmark & \checkmark & 44.06 & 52.30 & 0.13 & 0.05 \\
 & \checkmark & GT & \checkmark & 45.06 & 68.54 & 0.29 & 0.12 \\
\bottomrule
\end{tabular}
\vspace{-0.2cm}
\end{wraptable} 
% We compare~\cite{sadhu2021vidsitu} with our model on each of these settings.
% As clear from the table, our model beats VidSitu \cite{sadhu2021vidsitu} on all the settings on the task of SRL.
% discussed this in the verb accuracy now
% On Verb prediction, Our models performs slightly worse on accuracy than VidSitu, but performs better on macro-averaged recall, which reflects performance on all verb senses, instead of focusing on dominant classes.
% Coming to role predictions, VidSitu (GT+Pred) performs the best, since the model is trained using ground truth verbs which can directly predict the roles given the fixed verb-roles mapping.

% For (pred-pred) in VidSitu during SRL prediction we use the predicted verbs to infer on SRL. Vidsitu does slightly better than us on (pred-pred), for the same reason that, it was trained on ground truth verbs, which can easily learn to map verb-role pairs. While in our case we directly predict the roles from the contextualised event embeddings, circumventing the need for verb. All the above variants follow the same trend, i.e. as role prediction accuracy decreases so does the SRL captioning. 

\mypara{GVSR.}
We evaluate our end-to-end model for grounded video situation recognition.
In order to enable end-to-end GVSR in~\cite{sadhu2021vidsitu}, we use it in the ``Pred, Pred'' setting discussed above, that allows verb, role, and SRL predictions.
Table~\ref{tab:table7} shows that our model improves SRL performance over Vidsitu~\cite{sadhu2021vidsitu} by a margin of 22\% on CIDEr score.
In addition to that, our model also enables Grounded SRL, not achievable in VidSitu~\cite{sadhu2021vidsitu}.
% Due to space restrictions, we will include qualitative results in the Appendix.

\subsection{Qualitative Results and Limitations}
\begin{figure}[t]
\centering
\includegraphics[width=\linewidth]{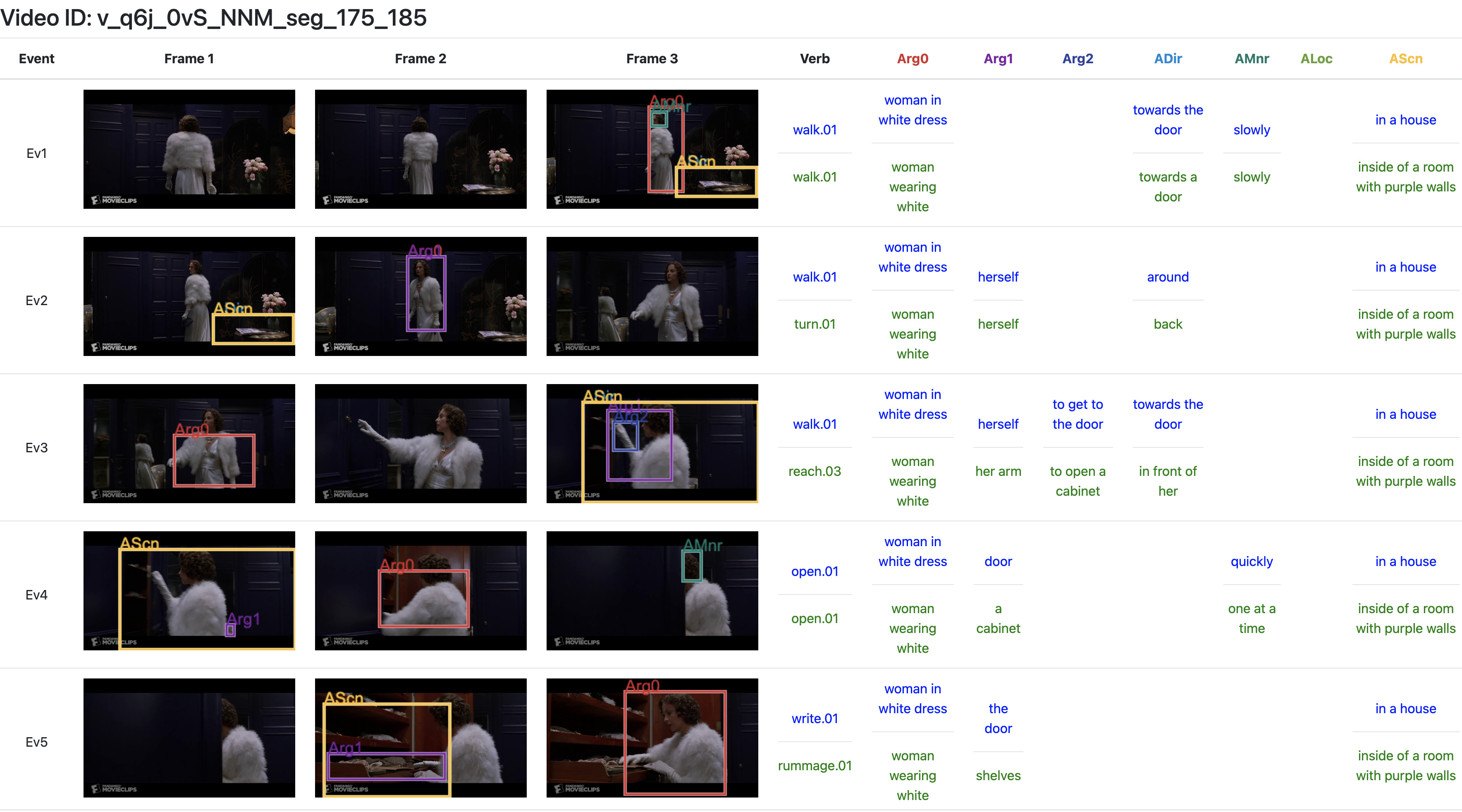}
\vspace{-0.6cm}
\caption{
We show the results for a 10s clip that can be viewed here: \url{https://youtu.be/q6j_0vS_NNM?t=175}.
The video is broken down to 5 events indicated by the row labels Ev1 to Ev5.
At a 1fps sampling rate, we obtain boxes from 3 frames for each event (with Frame3 of event $i-1$ being the same as Frame1 of event $i$).
On the right side of the table, we show the predictions for the verb and various roles in the ``Pred GT'' mode, discussed in Table~\ref{tab:table5} (row 6).
Predictions are depicted in blue, while the ground-truth is in green.
Each role is assigned a specific color (see table header), and boxes for many of them can be found overlaid on the video frames (with the same edge color).}
\vspace{-0.3cm}
\label{fig:qualitative1}
\end{figure}

We visualize the predictions of \modelname{} (Pred-GT) in Fig.~\ref{fig:qualitative1} for one video of 10 seconds%
\footnote{More examples on our project page, \url{https://zeeshank95.github.io/grvidsitu/GVSR.html}.}
and see that it performs reasonably well given the complexity of the task.
\modelname{} correctly predicts verbs for actions like ``open" and ``walk".
Given the large action space and complex scenes, there can be multiple correct actions, \eg~in Ev2 we see a reasonable ``walk" instead of ``turn''.

For SRL, the model generates diverse captions with good accuracy, like ``woman in white dress".
Even though the ground-truth is syntactically different, ``woman wearing white", they both mean the same.
In fact, this is our primary motivation to introduce grounding.
In Ev3, the model incorrectly predicts ``walk" as the verb instead of "reach".
While ``walk'' does not have the role Arg2, we are able to predict a valid caption ``to get to the door'' while grounding the woman's arm in Frame3.
We see that our model correctly understands the meaning of Arg2 as we use ground-truth role embeddings combined with event features for SRL.
This shows the importance of event embeddings, as they may recall fine-grained details about the original action even when there are errors in verb prediction.

For grounding SRL, we see that the model is able to localize the roles decently, without any bounding box supervision during training.
While we evaluate grounding only for Arg0, Arg1, and Arg2 (when available), we show the predictions for other roles as well.
In Fig.~\ref{fig:qualitative1}, the model is able to ground the visual roles Arg0 and Arg1 correctly.
For non-visual roles like \textit{``Manner"}, the model focuses its attention to the face, often relevant for most expressions and mannerisms.
% This shows that the model attends to relevant places while predicting the captions.

\mypara{Limitations}
for our current model are with verb and role prediction and disambiguation, improving the quality and diversity of captions to go beyond frequent words, and the division of attention towards multiple instances of the same object that appears throughout a video (details in Appendix~\ref{supp:sec:limitation}).
Nevertheless, we hope that this work inspires the community to couple videos and their descriptions.

%% file: sections/5_conclusion.tex
\vspace{-0.1cm}
\section{Conclusion}
\vspace{-0.1cm}
We proposed GVSR as a means for holistic video understanding combining situation recognition - recognizing salient actions, and their semantic role-noun pairs with grounding.
% , that allows answering several salient questions about a video that helps describing the entire situation with precision.
% This is step towards holistic video understanding
% covering action recognition, role-caption generation and localization.
We approached this challenging problem by proposing \modelname{}, that combines a video-object encoder for contextualised embeddings, video contextualised role query for better representing the roles without the need for ground-truth verbs and an event-aware cross-attention that helps identify the relevant nouns and ranks them to provide grounding.
We achieved state-of-the art performance on the VidSitu benchmark with large gains, and also enabled grounding for roles in a weakly-supervised manner.

%% file: sections/supplementary.tex
\newpage
\begin{center}
\textbf{{\large Grounded Video Situation Recognition - Appendix}}
\end{center}

We start with a brief mention of additional qualitative results from our model in Appendix~\ref{supp:sec:qualitative}.
Following this, we present quantitative results including an extended table that shows SRL performance on all VidSitu~\cite{sadhu2021vidsitu} metrics, and role prediction performance of our model (Appendix~\ref{supp:sec:quantitative}).
We also present challenges of predicting roles (which appear rarely) in Appendix~\ref{supp:subsec:role_prediction} and long-tail related challenges of predicting verbs in Appendix~\ref{sec:supp:longtail-verbs}.
We end this document by talking a bit about the limitations in Appendix~\ref{supp:sec:limitation} and the annotation process to obtain boxes on the validation set (Appendix~\ref{supp:sec:annotations}).

\section{More qualitative results}
\label{supp:sec:qualitative}

GVSR is a challenging problem that requires to correctly identify the action, disambiguate the roles taking part in it, localise the roles, and generate descriptive captions.
Moreover, the videos are curated from complicated movie scenes with fast motion, shot changes, and diverse scenes.
% We show results from the predicted-verb, and GT-role  \textit{("pred, GT")} version of VideoWhisperer.
For better visualisations we create an HTML file \href{https://zeeshank95.github.io/grvidsitu/GVSR.html}{GVSR.html}, with predictions on 10 videos.
There are a total of 5 events in each video.
As described in the method section, we sample 11 frames from the entire video, divided between 5 events, each with 3 frames with 1 frame sharing the event boundary.
Fig.~\ref{fig:qualitative1} of the main paper illustrates an example of this visualization.

\section{Additional quantitative results and metrics}
\label{supp:sec:quantitative}

\subsection{SRL Evaluation on the Test Set}
We evaluate our model on the test set from the evaluation servers of VidSitu~\cite{sadhu2021vidsitu}.
A constant improvement over~\cite{sadhu2021vidsitu} can be seen in Table~\ref{tab:table8}.
The trend is similar when compared with Table~\ref{tab:table4} from the main paper that reports performance on the validation set.

Note that we do not report grounding metrics as the ground-truth nouns are not available. We are currently working with the authors of VidSitu~\cite{sadhu2021vidsitu} to establish this as part of the benchmark.
% We are also currently unable to evaluate the verb predictions on the test set due to issues with the evaluation server.

\begin{table}[h!]
\begin{center}
\caption{Results of SRL with GT verb and role pairs on the test dataset. VidSitu's~\cite{sadhu2021vidsitu} results are as reported in their paper.}
\label{tab:table8}
\begin{tabular}{lccccccc}
\toprule
Method & CIDEr & C-Vb & C-Arg & R-L & Lea & IoU@0.3 & IoU@0.5 \\
\midrule
SlowFast+TxE+TxD~\cite{sadhu2021vidsitu} & 47.25 & 52.92  & 45.48 & 43.46 & {\bf 50.88} & - & - \\
\modelname{} (Ours) & {\bf 68.04} & {\bf 81.23} & {\bf 62.19} & {\bf 46.15} & 48.77 & - & - \\
\midrule
\textcolor{gray}{Human Level} & \textcolor{gray}{83.68} & \textcolor{gray}{87.78} & \textcolor{gray}{79.29} & \textcolor{gray}{40.04} & \textcolor{gray}{71.77} & - & - \\ 
\bottomrule
\end{tabular}
\end{center}
\end{table}

\subsection{GVSR, all metrics}
In Table~\ref{tab:table9} we show the results of end-to-end GVSR on all the metrics. We can see a clear improvement over the "pred, pred" VidSitu~\cite{sadhu2021vidsitu} on all the metrics for SRL.
Due to lack of space, we showed only the primary metrics in Table~\ref{tab:table7}.

\begin{table}[h!]
\begin{center}
\caption{GVSR: Results for end-to-end situation recognition.
Our model architecture is VO+RO+C.}
\label{tab:table9}
\scalebox{0.87}{
\begin{tabular}{l ccc c ccccc cc}
\toprule
\multirow{2}{*}{Model} & \multicolumn{3}{c}{Prediction} & Verb & \multirow{2}{*}{CIDEr} & \multirow{2}{*}{C-Vb} & \multirow{2}{*}{C-Arg} & \multirow{2}{*}{R-L} & \multirow{2}{*}{Lea} & \multicolumn{2}{c}{IoU} \\
 & Verb & Role & SRL & Acc@1 & & & & & & 0.3 & 0.5\\
\midrule
VidSitu~\cite{sadhu2021vidsitu} & \checkmark & \checkmark & \checkmark & 46.79 & 30.33 & 39.56 & 23.97 & 29.98 & 35.92 & - & - \\
\midrule
\multirow{2}{*}{\modelname{}} & \checkmark & \checkmark & \checkmark & 44.06 & 52.30 & 61.77 & 38.18 & 35.84 & 38.00 & 0.13 & 0.05 \\
 & \checkmark & GT & \checkmark & 45.06 & 68.23 & 74.15 & 61.79 & 45.58 & 48.22 & 0.27 & 0.15 \\
\bottomrule
\end{tabular}}
\end{center}
\end{table}

\subsection{Role prediction}
\label{supp:subsec:role_prediction}
Role prediction is critical for end-to-end GVSR. We analyse its performance for each role separately. As can be seen from Table~\ref{tab:table10} roles like \textit{Arg0, Arg1, Ascn, ADir, AMnr} which appears a lot more frequently than other roles in the dataset, have both high precision and recall, suggesting that role prediction can be done with a reasonably high accuracy directly from the video features.
Other roles that appears less frequently have a good precision but a very low recall, which is expected due to the long tail nature of roles.

\begin{table}[h!]
\begin{center}
\caption{Precision, Recall and, F1 score for role-prediction performance on all the role classes. Architecture is VO + RO + C in the ``Pred Pred" mode.}
\label{tab:table10}
\begin{tabular}{lc ccc}
\toprule
Method & Role-name & Precision & Recall & F1 \\
\midrule
\multirow{11}{*}{\modelname{}} & Arg0 & 0.90 & 0.97  & 0.93  \\
 & Arg1 & 0.79 & 0.93  & 0.86  \\
 & Arg2 & 0.55 & 0.26  & 0.36  \\
 & Arg3 & 0.30 & 0.05  & 0.09  \\
 & Arg4 & 0.15 & 0.04  & 0.06  \\
 & AScn & 0.74 & 0.93  &  0.83 \\
 & ADir & 0.66 & 0.49  & 0.56  \\
 & APrp & 0.36 & 0.03  & 0.06  \\
 & AMnr & 0.71 & 0.66  & 0.68  \\
 & ALoc & 0.40 & 0.12  & 0.19  \\
 & AGol & 0.65 & 0.15  & 0.24  \\
\bottomrule
\end{tabular}
\end{center}
\end{table}

\section{Long Tailed Verb Classification}
\label{sec:supp:longtail-verbs}
The grounded SRL task depends heavily on the action information.
In addition to complex scenes, the VidSitu dataset encompasses a large number of verbs and has a long-tailed distribution.
In fact, the number of verbs, 1560, is 2-4x larger than popular large-scale video action recognition datasets (Kinetics400 / Kinetics700).
We believe that these are the key challenges that result in lower performance for verb classification which inevitably affects the SRL.

We experiment with three common approaches to handle long-tailed distributions.
(i)~Loss re-weighting applies weights corresponding to the inverse verb frequency to the cross-entropy loss;
(ii)~Focal loss is applied as described in~\cite{Lin_2017_ICCV} (with gamma = 2.0); and
(iii)~Balanced sampling, we apply a weight for each sample such that the {\tt DataLoader} picks samples with a higher weight. The results are presented in Table~\ref{tab:table11}.

\begin{table}[h]
\begin{center}
\caption{Results of three common approaches to handle long-tailed distribution of verbs. V only represents the Video encoder (no object features) trained only for verb prediction.}
\label{tab:table11}
\begin{tabular}{lc}
\toprule
Method & Verb Acc@1 \\
\midrule
V only & 48.82 \\
\midrule
V only + Loss Re-weighting & 48.91 \\
V only + Focal loss & 47.81 \\
V only + Balanced sampling & 35.38 \\
\bottomrule
\end{tabular}
\end{center}
\end{table}    

Unfortunately, we do not see any significant improvement using these simple approaches. We have observed that the dataset is very challenging and has complex movie events with fast shot changes and many actions can be confusing. For example in Figure~\ref{fig:qualitative1} the woman turns while walking, but the model predicts ``Walk'' instead of ``Turn'' which is the dominant, but less significant action (if one considers duration).
Balanced sampling in particular leads to a significant drop since our sample consists of 5 event clips, each with a verb.
When rare verbs are oversampled, co-occurring event clips with potentially not-so-rare verbs are also oversampled, leading to a skewed training dataset.
This is similar to the challenges of applying balanced sampling to multi-label classification.

\section{Limitations}
\label{supp:sec:limitation}

Major challenges in GVSR include:
(i)~role disambiguation,
(ii)~descriptive caption generation, and
(iii)~localisation.
We describe each aspect in detail.

\textbf{Role disambiguation}
directly depends on the event features, since we use role queries contextualised by event embeddings. As described in Sec.~\ref{subsec:exp:gvsr}, event embeddings help in disambiguation of role even when the predicted action is incorrect.
But in many cases when the event embedding captures an action very far from the ground-truth, the role query gets updated based on the incorrect action and this hampers role disambiguation, in turn affecting the quality of SRL captioning and grounding.

\textbf{Descriptive captioning.}
We are able achieve descriptive captioning by exploiting object features.
Our model is able to predict difficult long-tailed entities like ``Monsters" and descriptive captions like ``Man in red towel", with high accuracy.
However, the presence of ``Man in black jacket" or ``AMnr: with a smile" is undeniably high.

\textbf{Localising roles}
in a weakly supervised manner is a very challenging task, it requires to disambiguate the roles and shift the attention to the right object out of a large pool of objects. Since the supervision comes from captions, which are descriptive and may refer to multiple attributes of an object, the attention is divided among many objects and it is difficult to get the most representative object with high probability.
Our model is able to ground the roles reasonably well, but leaves a lot of room for improvement.

\section{Annotations}
\label{supp:sec:annotations}

\begin{figure}[h]
\centering
\includegraphics[width=0.9\textwidth]{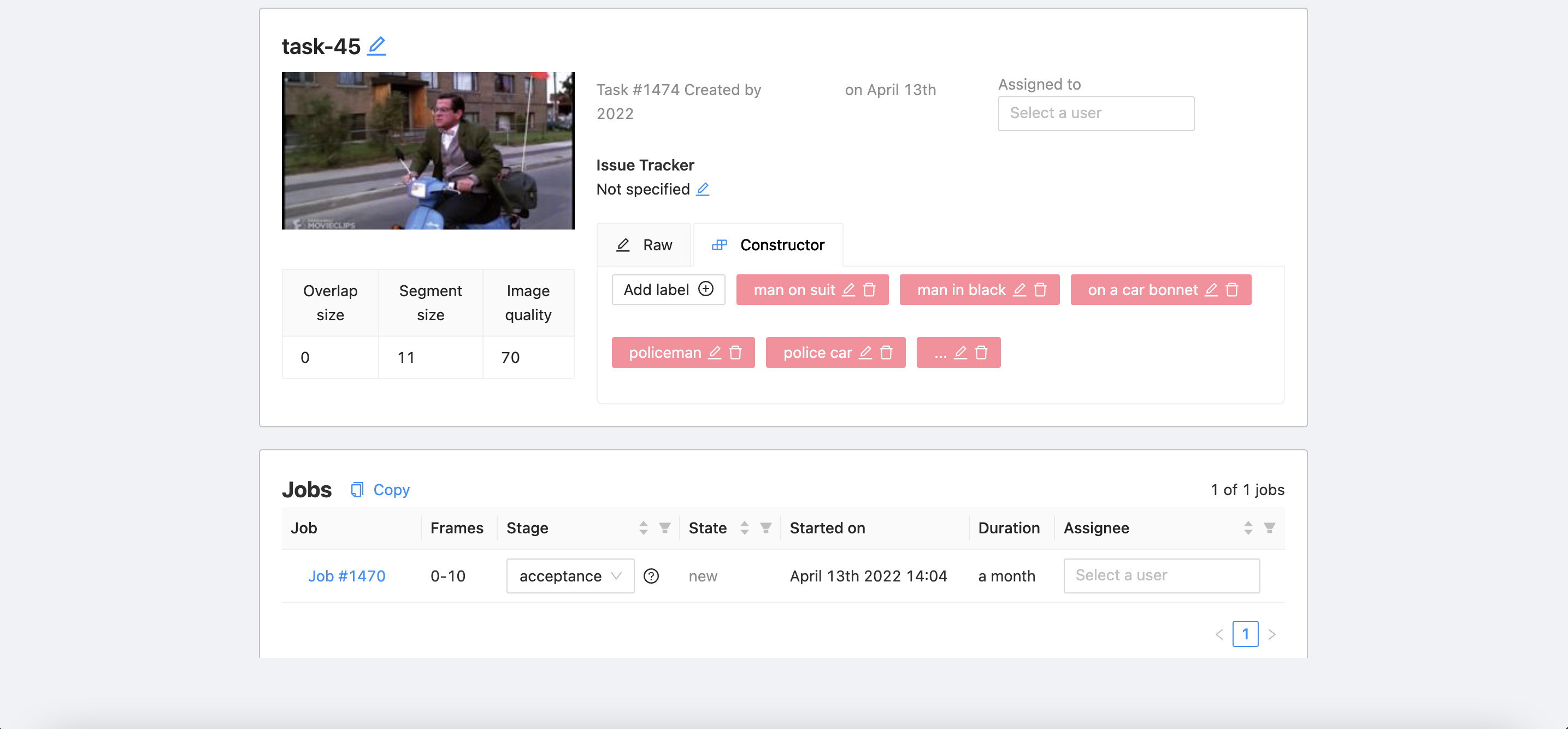}
\caption{Example annotation task for a video. There are a total of 11 frames subsampled at T=1 second from a 10 second video. Text highlighted in red are the labels.}
\label{fig:im1}
\end{figure}

\textbf{Sampling frames and creating an annotation task for a video.}
In a video of 10 seconds consisting of 5 events, we sample frames at 1fps, $\mcF = \{f_t\}_{t=1}^T$ from the entire video $V$, resulting in 11 frames.
Then, from the SRL annotations, we extract the captions for the typically visual roles: \textit{agent, patient, and instrument} from all the 5 events.
We retain all the unique captions from the selected ones and use them as ground-truth labels for the video $V$.
% Each video can have a different set of labels.
For each video we create a separate annotation task on the CVAT tool~\cite{cvat}, with video specific labels as shown in Fig~\ref{fig:im1}.

\subsection{Annotation Process}
\begin{figure}[h]
\centering
\includegraphics[width=0.9\textwidth]{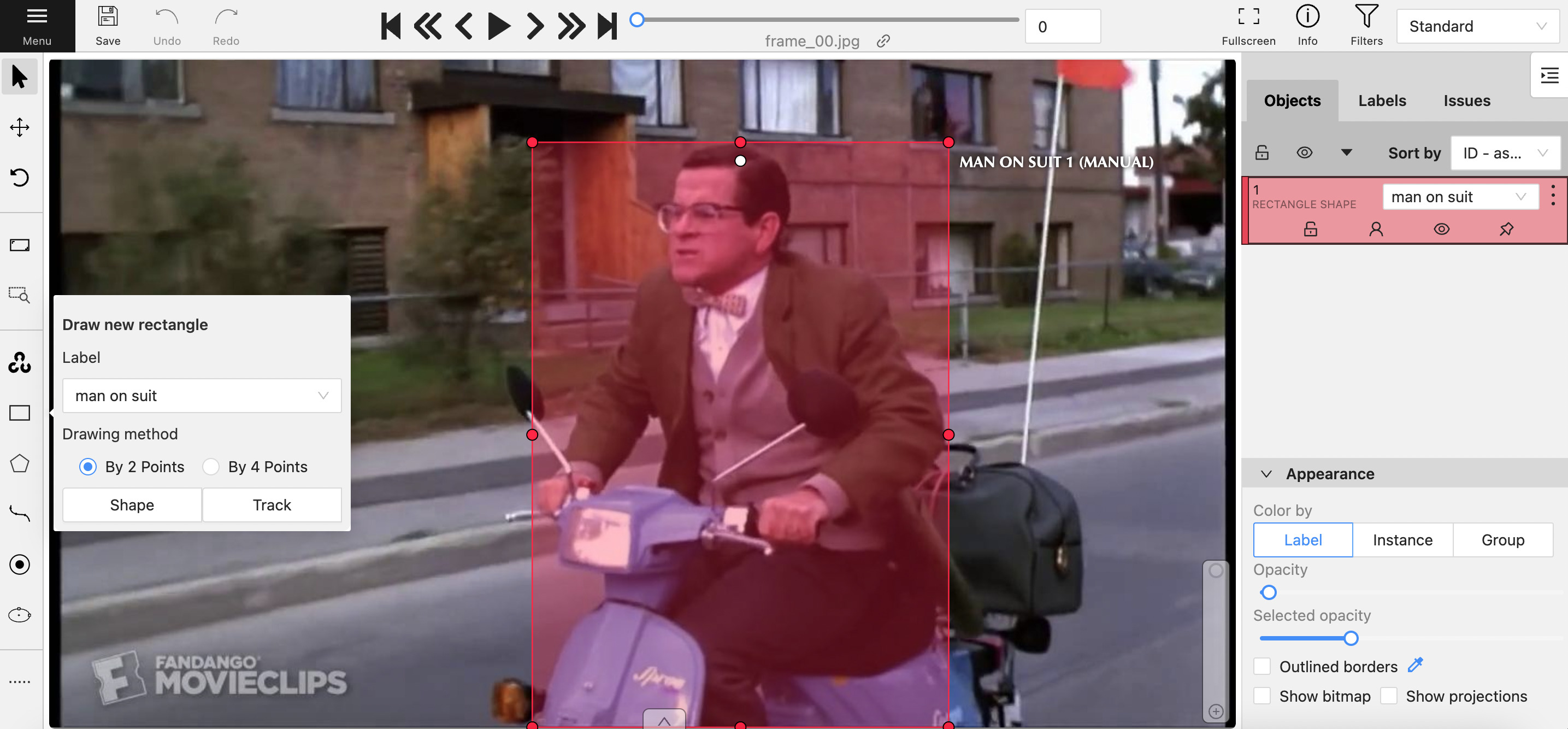}
\vspace{-0.3cm}
\caption{Select a label from the set of labels that can be visually recognised and draw a box around it.}
\label{fig:im0}
\end{figure}

\begin{figure}[h]
\centering
\includegraphics[width=0.9\textwidth]{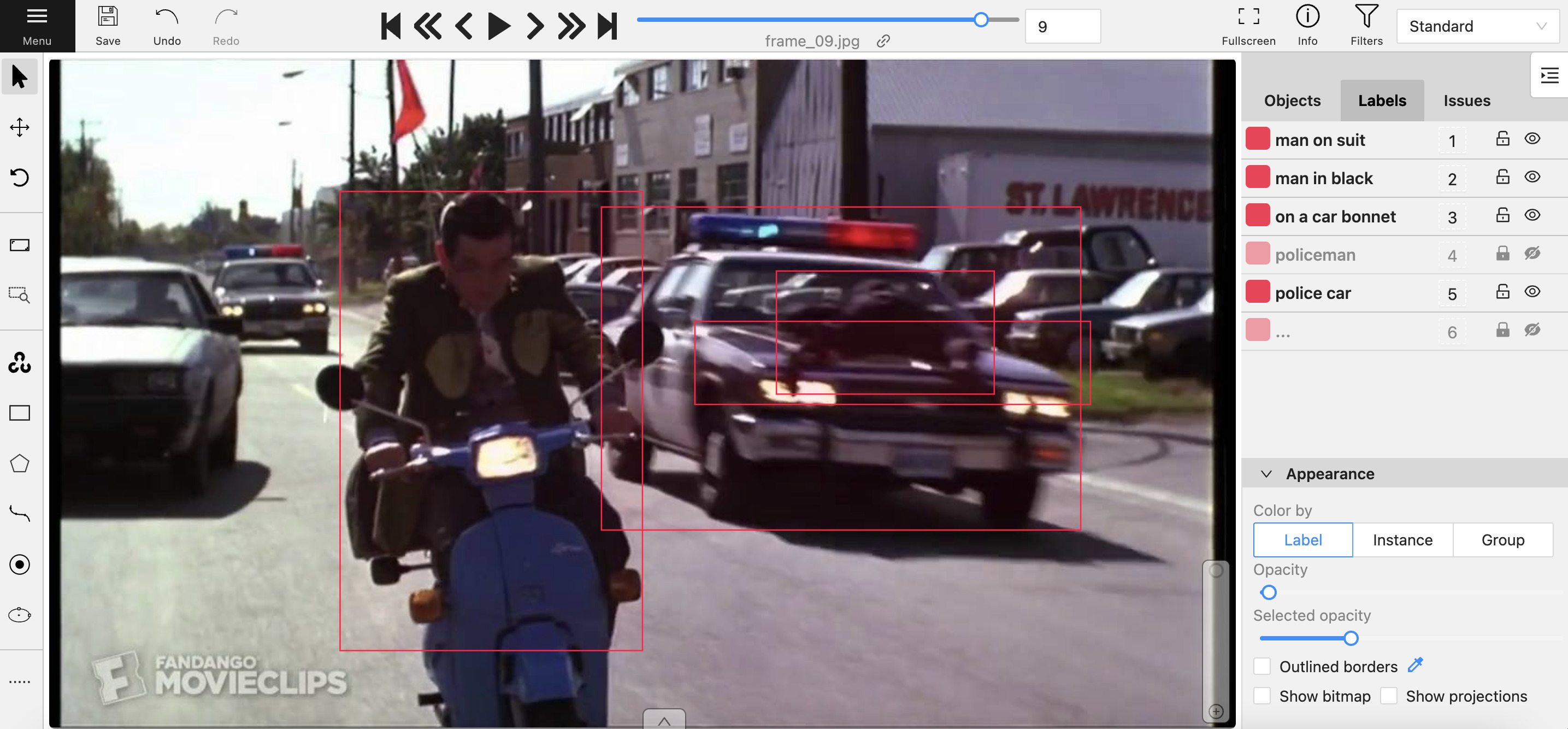}
\caption{Labels \textit{Man on suit, Man in black, on a car bonnet and, Police car} are visible in \textit{frame\_09}. Four boxes are drawn around the corresponding four entities}
\label{fig:im3}
\end{figure}

\begin{figure}[h]
\centering
\includegraphics[width=0.9\textwidth]{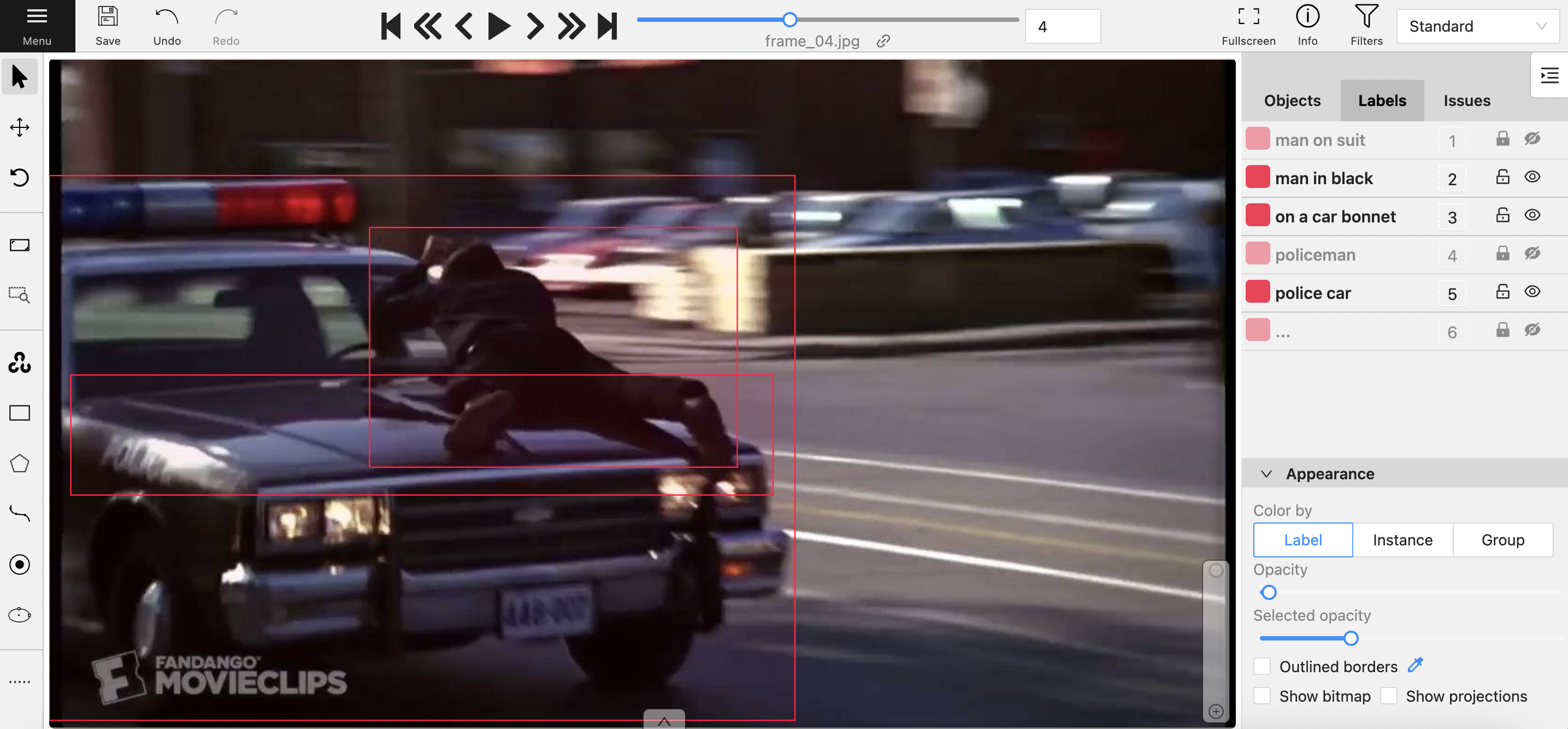}
\caption{Labels \textit{Man in black, on a car bonnet and, Police car} are visible in \textit{frame\_04}}. Three boxes are drawn around the three corresponding objects.
\label{fig:im4}
\end{figure}

\begin{figure}[h]
\centering
\includegraphics[width=0.9\textwidth]{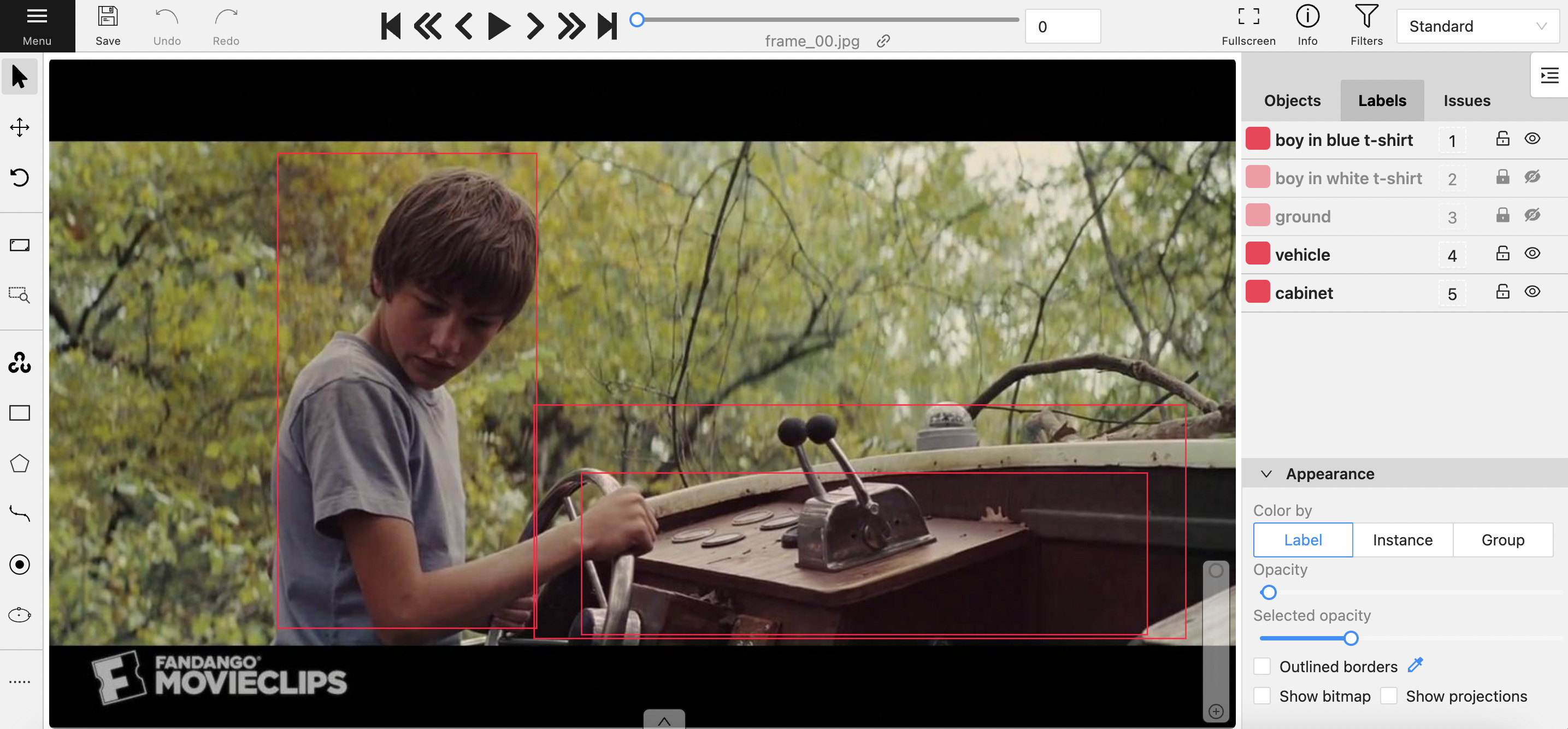}
\caption{Label \textit{ground} is not visible in \textit{frame\_00}, hence it is not annotated}
\label{fig:im5}
\end{figure}

\begin{figure}[h]
\centering
\includegraphics[width=0.9\textwidth]{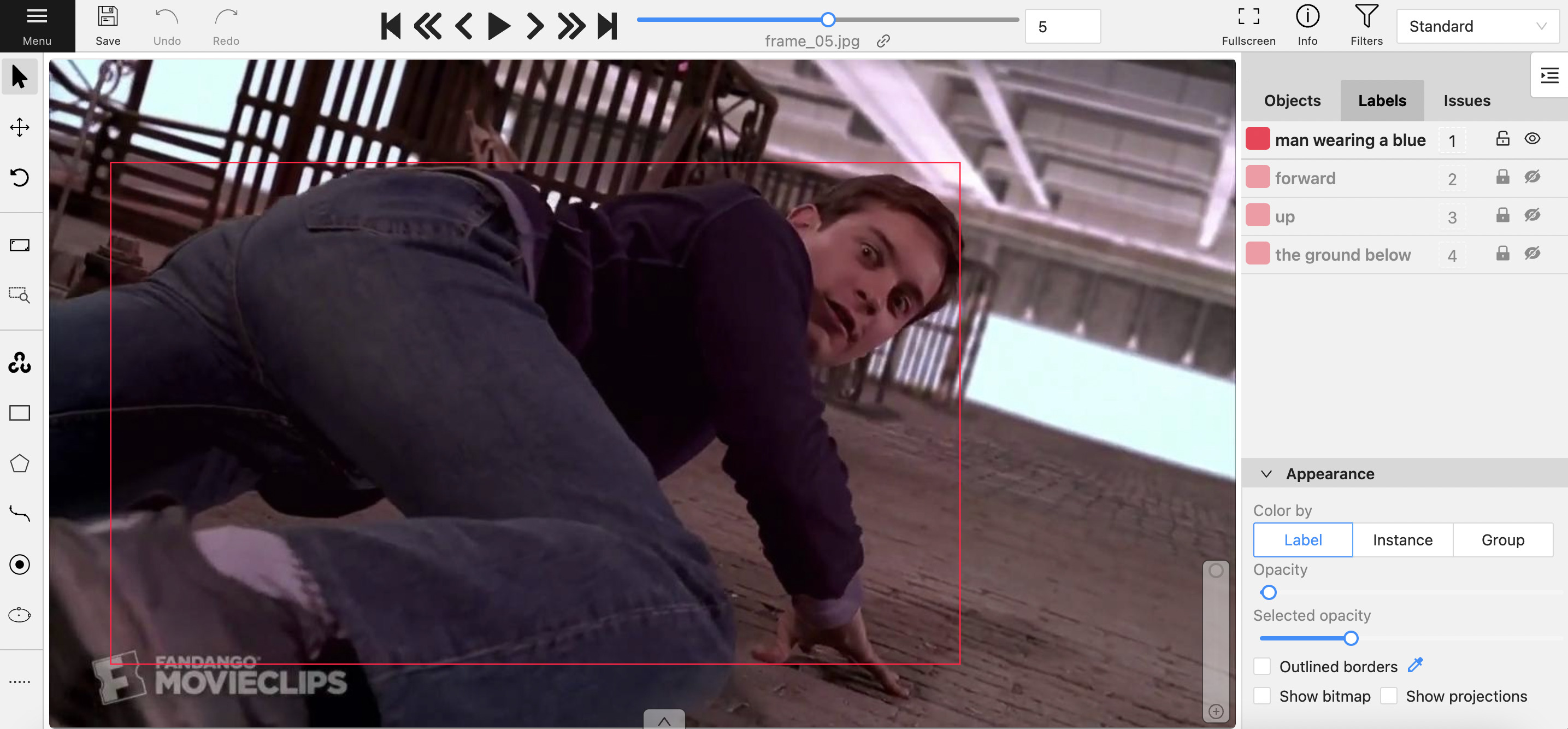}
\vspace{-0.3cm}
\caption{Label \textit{up} is a non-visual role, hence it is not annotated.}
\label{fig:im6}
\end{figure}

We iterate over every frame in $\mcF$, and find if any of the label is visually recognised.
If it is we select the label, and draw a bounding box around the visual entity as shown in Fig~\ref{fig:im0}, \ref{fig:im3}, and \ref{fig:im4}.
Some labels might not be visually present in the frames, like
\textit{Policeman} is not visible in any of the frames in Fig.~\ref{fig:im3} or
\textit{ground} is not visible in Fig~\ref{fig:im5}.
Some entities are non-visual like \textit{up} in Fig.~\ref{fig:im6}.
We do not annotate boxes for such roles.

After the annotations are done, for each event $i$ and role $k$ in a video, we create a dictionary of annotations $\mcG_{ik}$ with keys as frame number of all the frames that has the role $k$ annotated in it and values as the coordinates of the bounding box corresponding to them.
We will share the annotations for further research on our project page.
% \texttt{GT\_grounding.json} with the supplementary material.

\mypara{Compensation.}
We fairly compensated the annotators for their efforts at almost twice the minimum daily wage.
% a rate of 400 Rupees per day, which is 
\clearpage

